\pdfoutput=1
\documentclass[Afour,sageh,times]{sagej}

\usepackage{moreverb,url}

\usepackage[colorlinks,bookmarksopen,bookmarksnumbered,citecolor=red,urlcolor=red]{hyperref}

\newcommand\BibTeX{{\rmfamily B\kern-.05em \textsc{i\kern-.025em b}\kern-.08em
T\kern-.1667em\lower.7ex\hbox{E}\kern-.125emX}}

\def\volumeyear{2019}

\usepackage{xspace}
\makeatletter
\DeclareRobustCommand\onedot{\futurelet\@let@token\@onedot}
\def\@onedot{\ifx\@let@token.\else.\null\fi\xspace}

\def\eg{\emph{e.g}\onedot} \def\Eg{\emph{E.g}\onedot}
\def\ie{\emph{i.e}\onedot}

\def\wrt{w.r.t\onedot}

\usepackage{flushend}
\usepackage{multicol}
\usepackage{color}
\usepackage{graphicx} % for pdf, bitmapped graphics files
\graphicspath{{experiments/plots/}{./}}
\usepackage{amsmath} % assumes amsmath package installed
\usepackage{amssymb}  % assumes amsmath package installed

\DeclareMathAlphabet{\mbf}{OT1}{ptm}{b}{n}
\newcommand{\mbs}[1]{{\boldsymbol{#1}}}

\newcommand{\reffig}[1]{Fig.~\ref{#1}}
\newcommand{\refeq}[1]{Eq.~(\ref{#1})}
\newcommand{\reftab}[1]{Table~\ref{#1}}
\newcommand{\refsec}[1]{Section~\ref{#1}}

\newcommand{\pose}[2]{\ensuremath{^{#1}\mbs{\!P}_{#2}}}
\newcommand{\velocity}[2]{\ensuremath{^{#1}\mbs{v}_{#2}}}

\definecolor{todo-red}{RGB}{200,12,12}
\definecolor{green4}{RGB}{0,128,0}
\definecolor{blue4}{RGB}{0,0,128}

\begin{document}

\runninghead{Lynen, 
Zeisl, Aiger, Bosse,
Hesch, Pollefeys, Siegwart and Sattler}

\title{Large-scale, real-time visual-inertial localization revisited}

\author{Simon Lynen\affilnum{1,3},
Bernhard Zeisl\affilnum{1},
Dror Aiger\affilnum{2},
Michael Bosse, Joel Hesch,
Marc Pollefeys\affilnum{4,5},
Roland Siegwart\affilnum{3} and
Torsten Sattler\affilnum{6}}

\affiliation{\affilnum{1}Google Switzerland, Zurich\\
\affilnum{2}Google Israel, Tel Aviv\\
\affilnum{3}Autonomous Systems Lab, ETH Zurich\\
\affilnum{4}Computer Vision and Geometry Group, Department of Computer Science, ETH Zurich\\
\affilnum{5}Microsoft\\
\affilnum{6}Department of Electrical Engineering, Chalmers University of Technology (work done while at ETH Zurich)}

\corrauth{Simon Lynen,
Google Switzerland, Zurich,
Brandschenke Strasse 110,
8002 Zurich, Switzerland}
\email{slynen@google.com}

\begin{abstract}
%%%%%%%%%%%%%%%%%%%%%%%%%%%%%%%%%%%%%%%%%%%%%%%%%%%%%%%%%%%%%%%%%%%%%%%%%%%%%%%%
%2345678901234567890123456789012345678901234567890123456789012345678901234567890
%        1         2         3         4         5         6         7         8

%------------------------------------------------------------------------------

The overarching goals in image-based localization are scale, robustness and speed.
In recent years, approaches based on local features and sparse 3D point-cloud models have both dominated the benchmarks and seen successful real-world deployment.
They enable applications ranging from robot navigation, autonomous driving, virtual and augmented reality to device geo-localization.
Recently end-to-end learned localization approaches have been proposed which show promising results on small scale datasets.
However the positioning accuracy, scalability, latency and compute \& storage requirements of these approaches remain open challenges.
%End-to-end learned approaches also typically require encoding the geometry of the environment in the model, which causes performance problems in large scale scenes and results in a hard to accomodate memory footprint.
We aim to deploy localization at global-scale where one thus relies on methods using local features and sparse 3D models.
Our approach spans from offline model building to real-time client-side pose fusion.
The system compresses appearance and geometry of the scene for efficient model storage and lookup leading to scalability beyond what what has been previously demonstrated.
It allows for low-latency localization queries and efficient fusion run in real-time on mobile platforms by combining server-side localization with real-time visual-inertial-based camera pose tracking. 
In order to further improve efficiency we leverage a combination of priors, nearest neighbor search, geometric match culling and a cascaded pose candidate refinement step.
This combination outperforms previous approaches when working with large scale models and allows deployment at unprecedented scale.
We demonstrate the effectiveness of our approach on a proof-of-concept system localizing 2.5 million images against models from four cities in different regions on the world achieving query latencies in the 200ms range.
%\note{Torsten: It would be good to mention that our system is both more efficient than existing methods in the literature while handling a much larger scene. To the best of my knowledge, no other published system has been evaluated on a dataset of this scale.}
%\note{Torsten: What is important, but so far completely missing, is an explanation how our paper is related to robotics. Given that we are trying to publish in a robotics journal, I think this would be very important.}

\end{abstract}


\maketitle

%%%%%%%%%%%%%%%%%%%%%%%%%%%%%%%%%%%%%%%%%%%%%%%%%%%%%%%%%%%%%%%%%%%%%%%%%%%%%%%%
%2345678901234567890123456789012345678901234567890123456789012345678901234567890
%        1         2         3         4         5         6         7         8
%%%%%%%%%%%%%%%%%%%%%%%%%%%%%%%%%%%%%%%%%%%%%%%%%%%%%%%%%%%%%%%%%%%%%%%%%%%%%%%%
\section{Introduction}
\label{sec:chap8:introduction}
Visual localization, \ie, estimating the position and orientation of a camera in a given scene, is a fundamental problem in both Robotics and Computer Vision: 
Visual localization allows intelligent systems such as self-driving cars \citep{Haene2017IMAVIS} and drones \citep{Lim2015IJRR} to determine their current pose in an environment and thus to navigate to their target place. 
Localization enables SLAM systems \citep{Williams:etal:ICCV2007,mur2017orb} to detect and handle loop closure events.
It is a key building block of intelligent augmentation systems such as Augmented and Mixed Reality applications \citep{klein2007parallel,Castle:etal:ICRA2007,middelberg2014scalable}. 
Furthermore, visual localization is an important component of Structure-from-Motion (SfM) \citep{Schoenberger2016CVPR} systems.

Traditionally, visual localization algorithms rely on 3D maps and local features \citep{li2010location,li2012ECCV,Lim2015IJRR,sattler2017efficient,Zeisl2015ICCV,Svarm2017PAMI,Liu2017ICCV}.
They represent the scene as a 3D point cloud generated by SLAM or SfM, where each 3D point is associated with the local image descriptors from which it was observed. 
After extracting local features in an image taken by a camera, 2D-3D correspondences are established via matching the descriptors associated with the 2D query and 3D model points. 
These matches in turn are used to estimate the full 6 degree-of-freedom (DoF) pose of the camera, \ie, its position and orientation. 
Such structure-based methods have been shown to provide accurate pose estimates \citep{Walch2017ICCV} and scale up to a city-level \citep{li2012ECCV,Zeisl2015ICCV,Svarm2017PAMI}. 
In addition, real-time localization on mobile devices with restricted computational capabilities \citep{middelberg2014scalable,Lim2015IJRR,Lynen:RSS2015} is made possible through tightly integrating localization and SLAM algorithms. 

Feature-based methods consist of multiple stages, \ie, efficient descriptor matching, outlier filtering, robust camera pose estimation, and camera pose refinement. 
All of these stages directly impact both localization efficiency and accuracy. 
Implementing a high-quality localization system thus is a non-trivial task. 
Consequently, recent work has proposed to use convolutional neural networks (CNNs) to simplify the implementation by either learning the descriptor matching stage \citep{Donoser2014CVPR,Brachmann2017CVPR,Brachmann2018CVPR} or the complete localization pipeline \citep{kendall2015posenet,Kendall2017CVPR,Walch2017ICCV,Valada2018ICRA}. 
However, these approaches have been shown to be either inaccurate \citep{Walch2017ICCV,Sattler2017CVPR} or to not scale to larger or more complex scenes \citep{Brachmann2017CVPR,Taira2018CVPR,sattler2018benchmarking,Schoenberger2018CVPR}. 
In addition, they require a time-consuming training step every time a new scene needs to be considered or something changes in a scene. 
As such, classical feature-based methods are still highly relevant, albeit requiring careful design, algorithm and parameter choices.

This paper presents a feature-based visual localization system that is able to run in real-time on mobile platforms by combining server-side localization with real-time visual-inertial-based camera pose tracking. 
Using map compression to reduce memory requirements, our approach is able to scale to very large scenes spanning entire cities across the globe. 

In the conference paper version of this work \citep{Lynen:RSS2015} we focused on running localization on the device using highly compressed models that were transferred from the server.
One key contribution was a method for fusing the localization signal into the local state estimator such that it provided precise global tracking despite strong artifacts from model compression.
While this client-based approach scales to environments of a few 10,000 square meters it doesn't allow for the scale of deployment we are targeting today that spans city and country sized areas.
For city-scale localization we thus transitioned to an approach similar to~\citep{middelberg2014scalable} where server-side localization and client-side pose fusion are combined.
We base the system on improved and more scalable versions of the localization and pose-fusion algorithms proposed in \citet{Lynen:RSS2015} to allow for lower latency and higher pose accuracy than previous methods.

Given the complexities involved in implementing such a system, this paper aims at making possible design choices more transparent by discussing and evaluating multiple alternatives for each part of the pipeline.

%------------------------------------------------------------------------------
% Contributions.
%
In detail, this journal version makes the following contributions beyond our previous conference paper \citep{Lynen:RSS2015}:
\begin{itemize}
	\item We introduce and describe a system for visual localization suitable for large-scale server-side deployment.
	\item We provide a detailed discussion of all crucial parts, explain our design choices and point out alternative approaches. 
	%\item We describe how the results of the localization system is integrated into a visual-inertial pose tracking framework to allow high-precision real-time operation on mobile systems.
	\item We show how compression of the model requires adaptations of other parts of the localization system in order to achieve high localization performance.
	\item We extensively evaluate our approach by measuring the impact of each component and choice on run-time efficiency, pose estimation accuracy, and memory consumption.
	\item We evaluate our system at unprecedented scale both in terms of covered area and number of queries.
\end{itemize}

%------------------------------------------------------------------------------
%%%%%%%%%%%%%%%%%%%%%%%%%%%%%%%%%%%%%%%%%%%%%%%%%%%%%%%%%%%%%%%%%%%%%%%%%%%%%%%%
\section{Related Work}
On the topic of visual localization, \citet{piasco2018survey} provide an excellent and broad overview on the field. 
In the following, we thus focus only on the work that directly relates to our problem setup.

We review related work for all individual stages of our approach, \ie, model compression, feature-based localization, outlier filtering in visual localization, and combining global localization with local camera pose tracking. 
In addition, we review work on learning-based visual localization as well as work on place recognition, a problem closely related to the localization task.

\paragraph{3D map compression for visual localization:}
Given a 3D map reconstructed from a set of database images, where each 3D point is associated with one or more local feature descriptors, there are two basic approaches to map compression: 
Selecting a subset of points \citep{Cao2014CVPR,li2010location,park20133D}, thus compressing the 3D structure of the scene, or compressing the descriptors of the 3D points \citep{irschara2009structure,li2010location,Liu2017ICCV,sattler2017efficient,Sattler2015ICCV}.

Approaches of the first type select a minimal subset of 3D points such that all database images observe at least $k$ selected points, potentially while taking descriptor distinctiveness into account as done by \citet{Cao2014CVPR} and more recently by \citet{berrio2018identifying,van2018efficient, camposeco2018hybrid}. 
The underlying assumption is that the database images provide a reasonable approximation to the set of all probable viewpoints in the scene.
Approaches of the second type either represent each point via a subset of its descriptors \citep{irschara2009structure,li2010location,sattler2017efficient} or use vector-quantization to compress the descriptors \citep{Liu2017ICCV,Sattler2015ICCV}.

In this paper, we use both 3D structure and descriptor compression in a process we refer to as {\em summarization}:
specifically we sub-select 3D points, summarize the appearance and apply compression to the descriptors. 
Previous work focused on localizing individual images. 
This limited the amount of compression possible without negatively impacting localization accuracy. 
In contrast, our approach aims at localizing a moving camera and we show that this allows us to compress both appearance and geometry with little impact on the localization quality.

\paragraph{Feature-based localization:} 
Feature-based localization approaches use local patch descriptors \citep{alahi2012freak,Lowe:IJCV2004} to establish 2D-3D matches between 2D features extracted in a query image and 3D points in the map. 
These 2D-3D correspondences are then used to estimate the camera pose of the query image. 
This is typically done by applying a perspective-n-point-pose (PnP) solver, \eg, the 3-point-pose (P3P) solver for calibrated cameras \citep{Haralick1994IJCV}, inside a RANSAC \citep{Fischler:Bolles:ACM1981} loop.
Research on feature-based localization mainly focuses on efficient descriptor  matching \citep{Choudhary2012ECCV,Donoser2014CVPR,li2010location,Lim2015IJRR,sattler2017efficient} and outlier filtering for large-scale localization \citep{li2012ECCV,Liu2017ICCV,Zeisl2015ICCV,Sattler2015ICCV,sattler2017efficient,Svarm2017PAMI}. 
A popular approach to accelerate the correspondence search stage is to use prioritized matching \citep{Choudhary2012ECCV,li2010location,sattler2017efficient}. 
The underlying idea is that not all matches that are found end up being necessary for accurate camera pose estimation. 
Such approaches define prioritization functions for local features (in the case of 2D-to-3D matching, where features are matched against the 3D points) \citep{sattler2017efficient} or points (in the case of 3D-to-2D matching) \citep{Choudhary2012ECCV,li2010location}. 
Features / points are then matched in descending order of their priorities and correspondence search is terminated once a pre-defined number of matches is found. 
Rather than using prioritization, \citet{Donoser2014CVPR} model the 2D-3D matching stage as a classification problem. 
They train random ferns to efficiently determine the corresponding 3D point for each feature. 
However, the increase in efficiency comes at the price of matching quality. 
They thus need a pose prior, \eg, from GPS, to find sufficiently many correct matches.
To accelerate matching on mobile devices, \citet{Lim2015IJRR} amortize correspondence search over multiple frames. 

\paragraph{Outlier filtering for large-scale localization:} 
The localization approaches discussed above assume that the local appearance of each 3D point is rather unique. 
However, this assumption is often violated at large scale as local descriptors become more ambiguous as the size of scenes grows \citep{li2012ECCV}. 
Uniqueness is further reduced when descriptors are compressed, \ie, in the setting considered in this paper. 
Scalable feature-based localization approaches thus typically relax the matching criteria, \eg, by relaxing matching thresholds \citep{Camposeco2017CVPR,li2012ECCV,Svarm2017PAMI,Zeisl2015ICCV} or using quantized descriptors \citep{Liu2017ICCV,Sattler2015ICCV}, and use outlier filtering techniques to detect and reject wrong correspondences.

There are two dominant approaches to outlier filtering: 
1) Co-visibility-based methods \citep{Alcantarilla2011ICRA,li2012ECCV,Liu2017ICCV,sattler2017efficient} use the fact that the 3D point clouds generated by SfM and SLAM also contain visibility information \citep{li2010location}. 
More precisely, such 3D maps encode which 3D points are observed together. 
This information is used to select a subset of matches that is more likely to be correct from a larger set of matches \citep{li2012ECCV,Liu2017ICCV,sattler2017efficient}. 
2) In contrast, geometric outlier filtering methods determine for each match how consistent it is with the other matches \citep{Camposeco2017CVPR,Larsson2016BMVC,svarm2014CVPR,Zeisl2015ICCV,Svarm2017PAMI}. 
Correspondences that are consistent with only a few other matches are then removed before camera pose estimation.

In this paper, we quantize the 3D point descriptors to reduce the memory footprint of 3D maps, allowing our approach to scale to very large scenes. 
We employ a variant of the geometric constraints used by \citet{Zeisl2015ICCV} for filtering outliers that make up a large fraction of the matches in our setup.

\paragraph{Place recognition:} 
Given a database of geo-tagged images, the place recognition problem asks to determine the place depicted in a query image \citep{Arandjelovic2016CVPR,lynen2017trajectory,sattler2016large,Torii2015CVPR,Torii2015PAMI}. 
It is usually modelled as an image retrieval problem \citep{sivic2003video}, where the geo-tag of the most similar database image is used to approximate \citep{Zamir2010ECCV} or compute \citep{Sattler2017CVPR,Zhang06TDPVT} the geo-tag of the query image. 
The place recognition problem is relevant for loop closure detection in SLAM \citep{cummins2011appearance,GalvezTRO12,lynen2017trajectory}. 

Place recognition techniques are also related to the visual localization problem as they can be used to determine which part of a scene might be visible in a query image \citep{Cao2013CVPR,irschara2009structure,Sattler2015ICCV}, thus restricting the search space for feature matching. 
As such, place recognition techniques are used to reduce the amount of data that has to be kept in RAM, as the regions visible in the retrieved images might be loaded from disk on demand \citep{arth2009wide,tran2018device}.
Yet, loading 3D points from disk results in high query latency.
Our approach thus does not use an intermediate image retrieval step.
Instead, it relies on efficient model compression and direct search in the model.

\paragraph{Learning-based localization:} 
Visual localization and place recognition approaches can benefit from machine learning by replacing hand-crafted descriptors and image representation with learned alternatives \citep{Arandjelovic2016CVPR,Balntas2016BMVC,Brown2011PAMI,Lepetit2006,Mishchuk2017NIPS,Radenovic2016ECCV,Schoenberger2017CVPR,Weyand2016ECCV}. 
More advanced learning-based approaches directly replace (parts of) the localization pipeline. 
There are two dominant approaches in the literature: 
Learning to predict scene coordinates \citep{Shotton13CVPR,Valentin15CVPR,Brachmann2017CVPR,Brachmann2018CVPR,Cavallari17CVPR} and learning to directly regress camera poses \citep{kendall2015posenet,Kendall2017CVPR,Walch2017ICCV,Valada2018ICRA}. 
The former type of methods learn the 2D-3D matching stage of localization systems by learning to predict a 3D point coordinate for each pixel in a query image. 
The resulting 2D-3D correspondences can then be used for traditional camera pose estimation by applying a PnP solver inside a RANSAC loop. 
The latter type of methods directly learn the full localization pipeline by learning to predict the 6DOF camera pose using just an image as input. 

Given powerful enough hardware, learning-based methods can be run at acceptable latency (on low-resolution images). 
At the same time, they represent the scene implicitly through weights stored in CNNs or random forests, resulting in representations of a few MB for a room sized environment. 
Results show that learning-based methods can outperform classical feature-based pipelines \citep{Brachmann2018CVPR,Valada2018ICRA}, especially in weakly textured scenes~\citep{Walch2017ICCV}. 
However, they also have been shown to struggle to adapt to larger and more complicated datasets \citep{Brachmann2018CVPR,sattler2018benchmarking,Schoenberger2018CVPR,Taira2018CVPR,Walch2017ICCV}. 
This paper focuses on proposing a pipeline for server-side localization in city scale scenes. 
Thus, our approach follows classical feature-based pipelines.

\paragraph{Combining global localization and local camera pose tracking:} 
Localizing a single image against a large map is often too computationally complex to be done in real-time, especially on autonomous systems with limited computational capabilities such as drones or mobile devices.
But also streaming video to a server typically would incur too high latencies.

Once an initial pose is estimated from localization, \citet{Lim2015IJRR} use the visibility information from the 3D map to predict which 3D points might be visible in the next frame. 
Coupled with tracking already visible points, this restriction in search space enables real-time processing and high quality registration.
In order to scale to larger scenes, \citet{middelberg2014scalable} and \citet{ventura2014global} decouple localization and camera pose tracking. 
Localization is performed on an external server and returns the inlier 2D-3D matches if pose estimation is successful. 
These correspondences are then integrated, either on the server~\citep{ventura2014global} or on the device~\citep{middelberg2014scalable}, into the bundle adjustment of a PTAM-like~\citep{klein2007parallel} real-time SLAM system. 
By fixing the positions of the 3D points in these matches, the resulting additional constraints effectively prevent drift during on-device camera pose tracking. 
In this paper, we follow the approach from \citet{middelberg2014scalable} with two differences: 
1) compression techniques allow for reduced model size and unlock scalability to unprecedented scale while efficient algorithms allow latencies in the 200ms range.
2) we integrate the 2D-3D matches from localization into a filter-based visual-inertial odometry system \citep{Mourikis:etal:TRO2009}. 
Similar to recent work on filter based integration of localization \citep{dutoit2017consistent,kasyanov2017keyframe} this leads to significantly faster processing times and smoother pose tracking results compared to the approach used by \citep{middelberg2014scalable}.

%------------------------------------------------------------------------------
%%%%%%%%%%%%%%%%%%%%%%%%%%%%%%%%%%%%%%%%%%%%%%%%%%%%%%%%%%%%%%%%%%%%%%%%%%%%%%%%
\section{Building a Localization System}
\label{sec:chap8:overview}

\begin{figure*}[ht]
% \begin{tabular}{c}
  \centering
\includegraphics[width=0.75\textwidth]{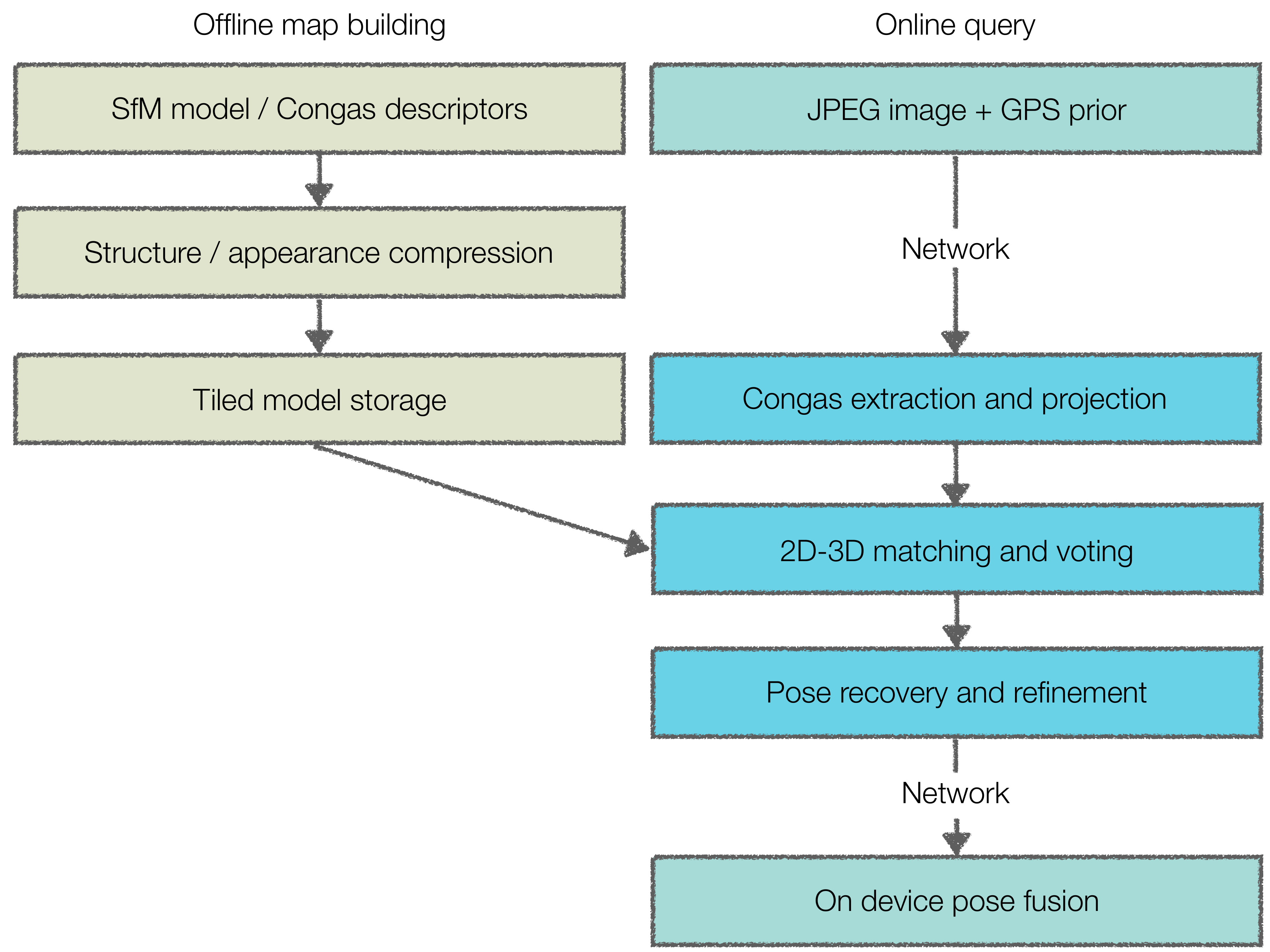}
% \end{tabular}
\caption{\small Overview of the complete system described in this paper, comprising of offline stages to create and compactly store the model of the environment and a real-time online query stack.
The online query part on the right is split between device and server with a mobile network connection link using about 20kB per VGA resolution query image.}
\label{fig:chap8:system_overview}
\end{figure*}

Fig.~\ref{fig:chap8:system_overview} illustrates our proposed localization approach. 
In the following, we briefly provide an overview over our system. 
We then describe each sub-component in more detail in subsequent sections. 

In an offline stage, a 3D point cloud is reconstructed from a set of database images using SfM.
In order to scale to large areas, we split this 3D point cloud model into subparts which cover a few street-blocks each.
Each of these sub-maps is compressed by summarizing the scene structure and appearance to reduce storage, compute cost and latency for the localization queries.

The robot or mobile-phone we want to geo-localize runs a real-time visual inertial SLAM method to track the movement of the camera in a {\em local} coordinate frame.
Local pose tracking provides the real-time signal for control or rendering and runs independent of the server using only on-device resources. 
For a subset of frames captured by the device camera, visual features are extracted and their descriptors are used to match against the descriptors of the previously built 3D point cloud.
This matching step takes place on the server, where a relevant subset of the model parts are selected based on approximate location from GPS/WiFi signals.
Once the models are loaded, efficient matching algorithms identify those 3D points in the model that denote the same objects as visible in the query image.
The resulting 2D-3D correspondences are first filtered based on geometric constraints and subsequently used to robustly estimate the {\em global} camera pose via RANSAC \citep{Fischler:Bolles:ACM1981}.
Given that localization has too high latency for real-time processing, we asynchronously feed the localization results into the local camera pose tracking, which continues to track the state between successful queries.
The pose computed for the first frame provides the initial position and orientation of the mobile system \wrt to the global model.
This pose thus {\em anchors} the {\em local} reference frame of the SLAM system to the {\em global} coordinates of the model.
For all subsequently localized frames, the inliers to the estimated pose are integrated into the state estimator of the SLAM system to provide additional constraints (besides the features tracked by the local SLAM).
These additional constraints continuously improve the alignment of the local trajectory against the global model and correct for any drift in the local tracking  
\citep{mourikis2009vision,bresson2013making,middelberg2014scalable,Lynen:RSS2015}.

%------------------------------------------------------------------------------
%%%%%%%%%%%%%%%%%%%%%%%%%%%%%%%%%%%%%%%%%%%%%%%%%%%%%%%%%%%%%%%%%%%%%%%%%%%%%%%%
\section{Data and evaluation metrics used in this paper}
\label{sec:chap8:experiments}

To facilitate reading the paper and relating algorithmic description with results we interleaved the chapters with experimental evaluation.
We thus first discuss the data and evaluation metrics that will be used throughout the following sections.

%%%%%%%%%%%%%%%%%%%%%%%%%%%%%%%
\subsection{Model Data Acquisition}
We evaluate the system on models built for the cities of Paris, Tokyo, Zurich and San Francisco to highlight parameter impact across appearance variations and scale.
3D models for each city are created from {\em Street View} collects over the last 10+ years using rigs comprising of 7 or 15 rolling-shutter cameras mounted on cars and backpacks.
The camera pose and structure computation additionally leverages input from GPS, LiDAR, odometry and inertial sensors~\citep{klingner2013street} to create a {\em global} model of the environment.
The model is subsequently subdivided into pieces of approximately 150x150m representing a processing unit for a given query that is convenient to store and load.
We apply the model compression scheme discussed in \refsec{sec:chap8:model_compression}.
By enforcing a minimum of 200 of the selected landmarks to be visible in each of the cameras we obtain weak guarantees on covering all parts of the environment.
A typical such model after compression consists of about 100,000 3D points and 200,000 to 500,000 visual descriptors.

\subsection{Evaluation Data Acquisition and Reference Pose Computation}
For each of the cities (see \reftab{tab:datasets}), we recorded evaluation sequences using consumer phones mimicking typical user behavior for augmented reality or robot navigation.
These sequences are recorded from the side-walk with the phone facing downwards in walking direction or across the street.
The resulting viewpoints thus exhibit substantial viewpoint differences from the views contained in the database as these were captured by cars on the street.
We obtain ground truth reference poses for the evaluation sequences by automatically establishing 2D-3D matches for each frame in the evaluation sequence against the 3D model.
These candidates are subsequently filtered using spatial verification taking into account the relative pose between cameras provided by SLAM~\citep{leutenegger2014keyframe}.
After filtering we associate the observations from the evaluation dataset with the model landmarks.
Using these constraints between local SLAM trajectory and global 3D model, we run a full-batch visual-inertial bundle adjustment to align the evaluation trajectory with the model.
After alignment, the poses of the evaluation trajectory typically exhibit sub-meter/sub-degree errors wrt. the model and are thus used as reference poses for evaluation.

\begin{table*}[]
\caption{The area/size of the models captured by Street View cars and the evaluation datasets used in this paper captured on foot using a mobile phone.}
\label{tab:datasets}
\centering
\begin{tabular}{| l | c | c | c | c | r  |}
   \hline
 &  Model area & Evaluation images & Evaluation sequence length  \\ \hline
Paris & $11$ km$^2$ & $147,776$ & $110$~km  \\
Tokyo & $148$ km$^2$ & $1,688,965$ & $2,036$~km  \\
Zurich & $30$ km$^2$ & $275,587$ & $186$~km  \\
San Francisco & $34$ km$^2$ & $460,981$ & $518$~km  \\
   \hline
 \end{tabular}
%\vspace{-0.6cm}
\end{table*}

Parameter studies were conducted using the model of Tokyo and by sampling 10\% of the evaluation data ($\sim$170,000 queries) unless indicated otherwise.

\subsection{Evaluation metric and approach}
We use {\em Average Precision} as the main metric in this paper rather than precision/recall at a given acceptance threshold.
Average precision is computed from the full precision-recall curve by taking the average of precision values at given recall $p(r)$: 
\begin{equation}
\operatorname{AveP} = \int_0^1 p(r)dr
\end{equation}

We find average precision a more useful metric to compare performance given it's independence of the chosen acceptance threshold (for instance inlier ratio of RANSAC) for localization responses.
We consider a localization result to be correct if it's position error is less than 3m and 10 degrees wrt. the reference pose.

\subsection{Best performing and reference method}
\label{sec:chap8:best_performing_variant}
For each of the plots we then only replaced the module under consideration while leaving the rest of the system the same.
Plots that are added throughout the text to illustrate properties of the subsystems are typically run on the best performing variant of the system (as described in detail later):

\begin{itemize}
	\item Model with observation count compression with a budget of 500k descriptors.
	\item Descriptors projected to 16D and product-quantized.
	\item Random Grids matcher using an absolute threshold.
	\item Keypoint orientation and GPS prior used during matching.
	\item 4D voting using a maximum of 4 neighbors per query keypoint.
	\item p2p + gravity solver for absolute pose.
	\item Refinement of matches after voting.
\end{itemize}

%------------------------------------------------------------------------------
%%%%%%%%%%%%%%%%%%%%%%%%%%%%%%%%%%%%%%%%%%%%%%%%%%%%%%%%%%%%%%%%%%%%%%%%%%%%%%%%
\section{Global 3D Model Creation}
\label{sec:chap8:model_creation}

Depending on the deployment environment of the system, global 3D localization models are typically generated from data captured with sensors carried by cars, pedestrians, or even users of the system.
We found that high quality results are most reliably obtained when using video sequences and fusing them into a single metric model of the environment \citep{schneider2018maplab}.
In a typical Robotics setup, each such video sequence consists of thousands of images and high-frequency data from an inertial-measurement unit (IMU).

Sparse visual feature descriptors detected in the images are matched to descriptors from nearby cameras to form feature tracks.
The 3D location of all tracks and the poses of the cameras are jointly estimated using metric SLAM techniques \citep{leutenegger2014keyframe,Mourikis:etal:TRO2009} to form self-consistent trajectories (including GPS, odometry and other signals if available).
Subsequently, the sequences are co-registered by localizing frames from one sequence against the 3D models of other sequences and batch-filtering the results to remove outliers.
We found the key ingredient for robust co-registration being {\em metric} SLAM algorithms.
These allow for efficient rejection of outlier matches by comparing relative transformations from SLAM with those computed from consecutive localizations as done by \citet{zach2010disambiguating} for images in SfM.
After a rigid alignment of the trajectories based on inlier matches, we run a pre-optimization of a reduced system \citep{johannsson2013temporally,nerurkarc:etal:RSS2013} with cross-trajectory constraints to eliminate low frequency errors from the trajectory.
As a final step a MAP optimization (visual inertial bundle-adjustment) including inter- and intra-trajectory constraints with the following objective function is solved:
\begin{equation}
J(\mbf x) := \underbrace{\displaystyle\sum_{i = 1}^{I}
\displaystyle\sum_{j \in \mathcal J(i)}
{\mbf e^{i, j}_\mathrm{r}}^T {\mbf W^{i, j}_\mathrm{r}} {\mbf
e^{i, j}_\mathrm{r}}}_{\mathrm{visual}}
+ \underbrace{\displaystyle\sum_{\vphantom{j \in \mathcal J(i)}i=1}^{I-1} {\mbf
e^{i}_\mathrm{s}}^T {\mbf W^i_\mathrm{s}} {\mbf e^{i}_\mathrm{s}}}_{\mathrm{inertial}} \enspace.
\end{equation}
Here, $i$ denotes the camera frame index and $j$ denotes the landmark index.
The set $\mathcal J(i)$ contains the indices of landmarks visible in the $i^{\mathrm{th}}$ frame (from the same or another trajectory).
Furthermore, $\mbf W^{i, j}_\mathrm{r}$ represents the inverse measurement covariance of the respective visual observation, and ${\mbf W^i_\mathrm{s}}$ the inverse covariance of the $i^{\mathrm{th}}$ IMU constraint.
Even for areas of the size of small cities, these optimization problems quickly exceed hundreds of thousands of camera frames with tens of millions of landmarks.
In order to keep the computational time limited, we subdivide the problem into spatially disjoint sub-areas which are iteratively refined and smoothed \citep{guo2016large,schneider2018maplab}.
See~\citep{klingner2013street} for further detail on techniques for globally optimizing a large number of trajectories to obtain models covering areas at country scale.
%
%In order to allow fast loading and efficient storage the final resulting model is subdivided into smaller pieces of 150x150m each containing only the information captured by images taken within that region.

%------------------------------------------------------------------------------
%%%%%%%%%%%%%%%%%%%%%%%%%%%%%%%
\section{Descriptor Extraction, Selection and Projection}
\label{sec:chap8:descriptors}

Due to the computational and power constraints of mobile platforms, typically cheap, yet less repeatable and discriminative visual feature descriptors are used in the SLAM frontends \citep{leutenegger2014keyframe,mur2017orb}.
Common choices are BRIEF \citep{calonder2010brief}, ORB \citep{rublee2011orb}, FREAK \citep{alahi2012freak} descriptors using FAST \citep{rosten2006machine} or DoG \citep{Lowe:IJCV2004} interest points that can be computed at 30+ FPS on mobile platforms.
It was shown that these descriptors typically provide sufficient discriminative power to reliably perform loop-closure, recovery and even multi-user experiences in room-scale setups~\citep{hartmann2013comparison}.

The applicability of BRIEF, ORB, FREAK and BRISK for large scale scenes however is limited \citep{sun2017dataset}, in particular when the query images exhibit substantial scale and viewpoint changes.
Descriptors such as SIFT \citep{Lowe:IJCV2004} and SURF \citep{Bay06surf} provide better performance under these conditions as well as in large scenes with strong visual aliasing.
A recent analysis \citep{Schoenberger2017CVPR} showed that classical descriptors also still outperform learned variants such as LIFT \citep{yi2016lift}.
Learned descriptors do not yet seem to generalize well to new scenes that look sufficiently different to the training data, yet one can expect to see the situation change in the near future.
Other learned descriptors like Delf \citep{noh2017large} work well in retrieval tasks where fewer descriptors are beneficial and precise keypoint locations are not critical given that typically only approximate geometric re-ranking is performed.
For localization however a high number of precisely located keypoints typically translates to a higher pose accuracy. 

Trading computational cost and recognition performance is driven by the use-case and the allowed complexity of the system.
Obvious options are using computationally cheap descriptors throughout the system (SLAM and global localization) or computing costlier descriptors periodically for localization queries only.
As argued in our previous publication \citep{Lynen:RSS2015}, picking a single descriptor for the entire stack allows for a direct integration of localization matches into the SLAM frontend and thus reduces system-complexity compared to having two descriptor types.
A separate high-quality descriptor on the other hand can significantly boost localization recall.

In the following, we focus on large scale scenes ($>20,000~m^2$) as found in indoor malls and transit stations as well as outdoor urban areas where GPS/WiFi is typically available, but has reduced accuracy. 
We explicitly do not focus at solving the problem of localizing {\em globally} or in an entire city since for all practical matters GPS/WiFi receivers are ubiquitous and limit the search radius.
To ensure reliable performance at this scale, we compute 40~dimensional real-valued descriptors using the `COmpact Normalized GAbor Sampling' (CONGAS) algorithm \citep{zheng2009tour}.
As in our earlier work \citep{Lynen:RSS2015,lynen2017trajectory}, we project the descriptors to a low-dimensional real-valued space to provide runtime improvements.
For the experiments in this paper we use Principal Component Analysis (PCA) due to its simplicity and sufficient performance~\citep{lynen2017trajectory}.
We reduce the descriptor to $16$~dimensions by removing the dimensions with the lowest signal-to-noise ratio to limit storage and runtime cost.

Another option to perform dimensionality reduction is to pick a projection matrix such that the \textit{$L_2$} distance between descriptors in the projected space matches the Likelihood Ratio Test (LRT) statistic \citep{bosse2009keypoint,lynen2017trajectory}.
The LRT is the hypothesis test that best separates matching from non-matching descriptors for a given maximum false-positive matching rate and thus allows including weak supervision when computing a projection matrix for dimensionality reduction.
Other, well performing options include learned non-linear dimensionality reduction techniques as we have discussed in more detail in our earlier work \citep{loquercio2017efficient}.

%------------------------------------------------------------------------------
%%%%%%%%%%%%%%%%%%%%%%%%%%%%%%%
\section{The Need for Model Compression}
\label{sec:chap8:model_compression}

When deploying a localization system, compactness for efficient storage as well as latency for loading and unpacking the 3D model from disk/network is a key factor.
In our previous publication \citep{Lynen:RSS2015}, our proposed compression provided compactness to allow on-device localization and efficient download of the model from the server.
However, also now that we want to run the localization system directly on server the same properties are key to scalability.
A typical model contains around $1M$ 3D points for an area of $20,000~m^2$ which without compression would require between $500$~MB and $800$~MB of storage. 
Thus, covering even a medium-sized city would quickly consume several tens of TB of storage.
Depending on the uncertainty of GPS, multiple sub-regions of the map need to be loaded to serve a given query and thus incur high latency.

Due to the density of data used to build the 3D models (videos or photo-collections with redundant views), only a subset of points from the initially reconstructed 3D model is necessary for successful localization.
This allows us to reduce the size of the model by a process we call {\em summarization}: a combination of 3D structure subselection and appearance summarization.
In addition to summarization we apply domain specific data compression to bring the storage requirements to $10$ to $15$~MB; a reduction of $>95\%$.

%------------------------------------------------------------------------------
%%%%%%%%%%%%%%%%%%%%%%%%%%%%%%%
\subsection{Structure Summarization}
\label{sec:model_summarization:structure}

When capturing a scene multiple times for model construction the system can incorporate information about changes in the environment and gain robustness.
At the same time however, redundant scene information leads to an increase in 3D model size.

For a real-world deployment where certain streets might have been captured hundreds of times, we thus need to solve a non-trivial subselection and compression problem:
Build and maintain a model of the environment which is constant in memory size, yet capable of incorporating newly collected data over time.
Early work in the Robotics community discarded entire image captures from the map~\citep{maddern2012towardsA,maddern2012towardsB} or formed full image descriptors~\citep{naseer2014robust} used to identify novel views of the scene.
Discarding entire images however ignores the fine grained information about stable environment features that are critical for robust localization.
While \citet{johns_icra2013} learn place dependent feature statistics to limit insertions to the database to relevant locations, the representation per location still grows unbounded.

Driven by the structure-based localization algorithm and the desire for a constant size model, we leverage subsampling and summarization techniques from the Computer Vision \citep{li2010location,park20133D,Cao2014CVPR,camposeco2018hybrid} and Robotics \citep{dymczyk2015gist,dymczyk2015keep,muhlfellner2016summary} communities.
These techniques have demonstrated that it's possible to discard large parts of the initial 3D points with only moderate influence on the localization performance. 

After such compression, only the minimal amount of data required for localization is retained, which are the 3D landmark positions together with their corresponding descriptors as well as landmark covisibility information.
Since the reduction process removes the landmarks which have been observed the least or have a large estimated covariance, the overall localization performance is only marginally impacted.
Performance declines more significantly only at high reduction rates \citep{li2010location}.
We have also observed high compression rates in our previous work \citep{Lynen:RSS2015,dymczyk2015gist}, but interestingly not on our city scale 3D models where the map density is typically lower than in SfM models built from photo tourism collections.

%The approaches \citep{li2010location, park20133D} select a minimal subset of all landmarks such that each cameras observes at least $N_\text{thres}$ (here 200) selected landmarks in order to target robust localization in all parts of the map.
%
Picking the optimal selection of landmarks is a set cover problem (NP-complete).

A simple heuristic implementation of this data reduction is to set an upper bound on the number of descriptors per area and assign a fraction of this total to each camera in the map; we denote this by `{\em observation count}' based heuristic.
For each camera one keeps landmarks that have been observed most, discarding others until the budget per camera is met.
Even given this simple strategy one can achieve good compression rates while ensuring coverage across the map.

While heuristics for the selection as used by \citet{li2010location} and \citet{dymczyk2015gist} provide good results, it has been shown by \citet{mauro2014ilp_cam_selection, havlena2013optimal_reduction} that optimization based approaches can provide gains especially under high compression rates.
These approaches target picking the most informative landmarks while ensuring that every camera must observe at least $N_\text{thres}$ landmarks to provide coverage across the model area.
This sampling process under a coverage constraints is a set-cover problem, which was shown to be NP complete~\citep{karp1972reducibility}.
Techniques such as Integer Linear Programming (ILP) have been used earlier by \citet{park20133D} and later by \citep{dymczyk2015keep, camposeco2018hybrid} to pick the optimal number of landmarks (here denoted by `{\em landmark count}' based).
We use the observation count of a landmark to score 3D points and then iteratively remove points from the model.

We apply our previous work~\citep{dymczyk2015keep} which extended the approach of \citet{park20133D} by augmenting the ILP problem with a term that constraints the absolute number of landmarks in the model.
Using this constraint allows bounding the total memory cost to a desired memory budget.
Adding this constraint to the problem however conflicts with the constraint of having a minimum number of observations for every camera and thus requires the addition of slack variables.
The optimization problem selecting the best landmarks under constraints is~\citep{dymczyk2015keep}:
\begin{equation} \label{eq:chap8:opt_landmarks_constraint}
\begin{split}
\text{minimize}\; \mathbf{q}^T\mathbf{x} + \lambda \mathbf{1}^T\boldsymbol{\zeta}\\ 
\text{subject to}\; \mathbf{A}\mathbf{x} + \boldsymbol{\zeta} &\geq b\mathbf{1} \\
\sum\limits_{i=1}^N \mathbf{x}_i &= n_\text{desired} \\
\mathbf{x} &\in \left\lbrace 0, 1 \right\rbrace^N \\
\boldsymbol{\zeta} &\in \lbrace\lbrace 0 \rbrace \cup \mathbb{Z}^+\rbrace^M.
\end{split}
\end{equation}
Here, $\mathbf{x}$ is a binary vector whose $i$th element is one if the $i$th point is kept in the model and zero otherwise.
$\mathbf{A}$ is an $M\times N$ visibility matrix where $M$ is the number of images and $N$ the number of points in the model.
$b$ denotes the minimum number of 2D-3D correspondences to keep for each camera and $\mathbf{q}$ is a weight vector which encodes the score (for instance observation count) per landmark.
The same variety of scoring functions as for the greedy approach from \citet{dymczyk2015gist} can be used with ILP by incorporating their results into the linear term weight $\mathbf{q}$.
While ILP based subselection is computationally expensive, we found it computationally tractable once the models are split into our 150x150m sized sub-maps and processed independently.

For each sub map we start from a full SfM model, comprising typically of 1000-2000 panoramic camera frames and up to $1M$ landmarks (before summarization).
We first investigate the difference between the {\em observation count} based heuristic (which compresses the map by removing observations) and the ILP-optimization based {\em landmark count} variant (which picks the optimal number of landmarks).
We create variants of the model for Tokyo with different thresholds for observation and landmark count and measure average precision over their memory footprint.

\begin{figure}[]
\begin{tabular}{c}
\includegraphics[width=0.98\columnwidth]{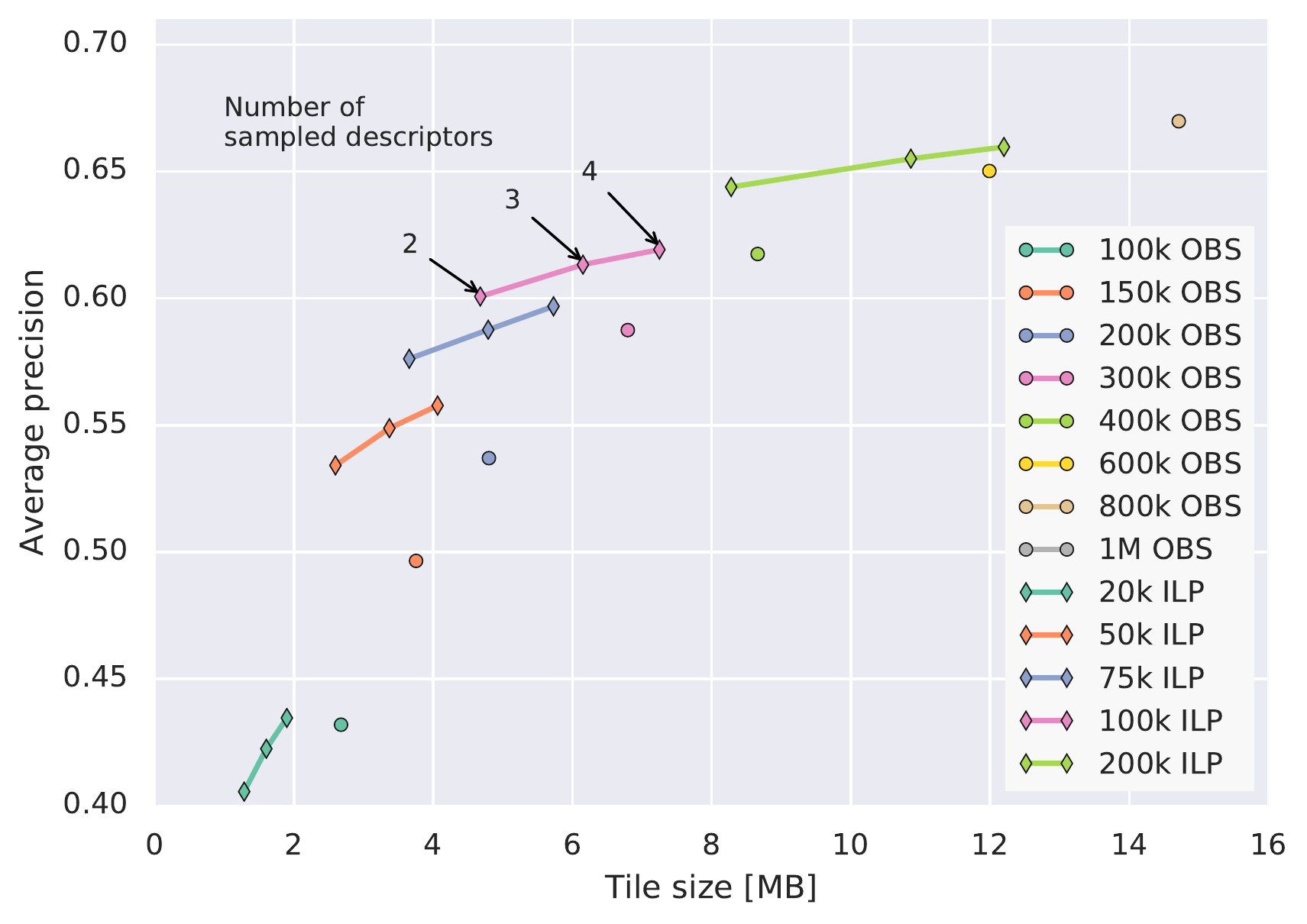}
\end{tabular}
\caption{\small Different compressed variants of the Tokyo model using a heuristic observation count selection (OBS)
 or a ILP optimization-based landmark selection (ILP) strategy.
 For the values of the OBS strategy, the numbers denote the budget of descriptors, while for the ILP strategy the numbers denote the budget of landmarks.
 We also plot different number of sub-selected descriptors for the ILP variant as reference.}
\label{fig:chap8:heur_vs_ilp}
\end{figure}

It is evident from \reffig{fig:chap8:heur_vs_ilp} that the ILP variant outperforms the heuristic independent of the compression rates for the descriptor for the same memory budget.

%------------------------------------------------------------------------------
%%%%%%%%%%%%%%%%%%%%%%%%%%%%%%%
\subsection{Appearance Summarization}
When performing map-compression based on ILP-optimization following \refeq{eq:chap8:opt_landmarks_constraint}, each resulting landmark is represented in the map by a set of descriptors.
Typically the different descriptors do not all represent unique modes of the landmark appearance and thus add substantial redundancy to the model.
Obvious choices for compressing the landmark appearance are to apply averaging or subselection as previously proposed by \citet{sattler2017efficient} (mean descriptor per visual word) and \citet{irschara2009structure} (k-medoid clustering to find a subset of relevant descriptors).

While the compression strategy based on landmarks provides a useful reduction, landmarks from densely collected areas have a high number of observations and thus consume much of the memory budget.
While the different observations of the landmarks capture the varying modes of appearance, they often exhibit redundancy and can be equally well represented with fewer modes.
Besides the structure compression by landmark selection we thus additionally perform per landmark {\em appearance summarization} to store only the relevant appearance modes per landmark.
We compare standard density analysis techniques with random subselection and averaging and show their tradeoffs wrt. memory and localization performance.

%For the {\em ILP-optimization based} variant we also added the different operating points for descriptor compression (picking 2, 3, 4 descriptors) as a reference, which is applied for the descriptors of selected landmarks.

\begin{figure}[]
\begin{tabular}{c}
\includegraphics[width=0.98\columnwidth]{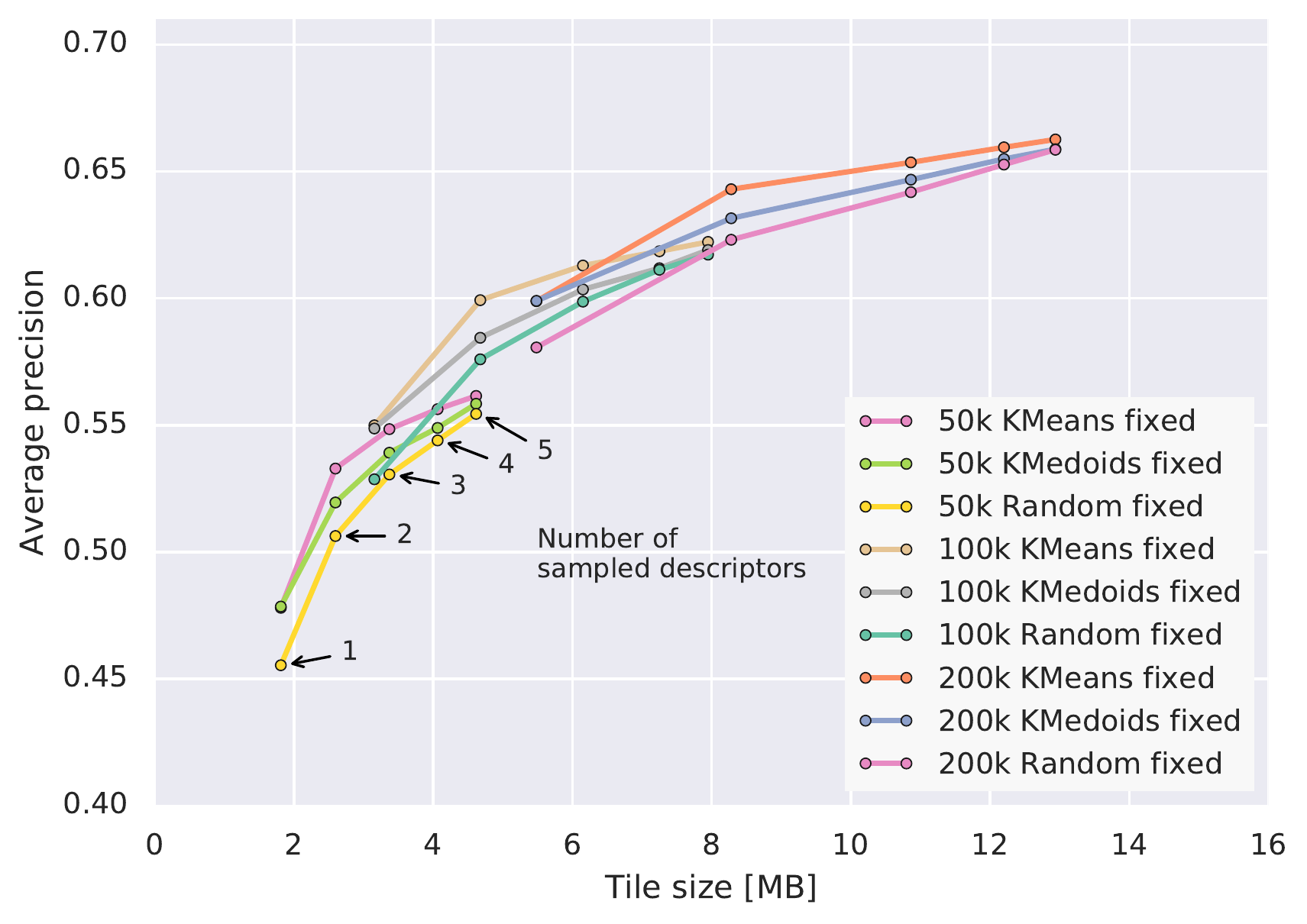}
\end{tabular}
\caption{\small Combining ILP optimization based landmark selection with appearance compression using a fixed number of target descriptors per landmark.
We compare performance on the Tokyo model using k-means or k-medoids centers with a random selection for different landmark budgets (denoted by the leading number).}
\label{fig:chap8:ilp_fixed}
\end{figure}

As shown in \reffig{fig:chap8:ilp_fixed}, it is possible to reduce the memory footprint by around 40\% at moderate performance loss.
Picking the centers per landmark from k-means clustering is superior to centers from k-medoids or random selection.

Given the difference in visual saliency for points in the environment, the number of observations varies between landmarks.
We found that observation count also correlates with the variance in appearance for landmark descriptors.
We thus also evaluate the performance of using a fraction of the original descriptor count as the target number of descriptors to pick.

\begin{figure}[]
\begin{tabular}{c}
\includegraphics[width=0.98\columnwidth]{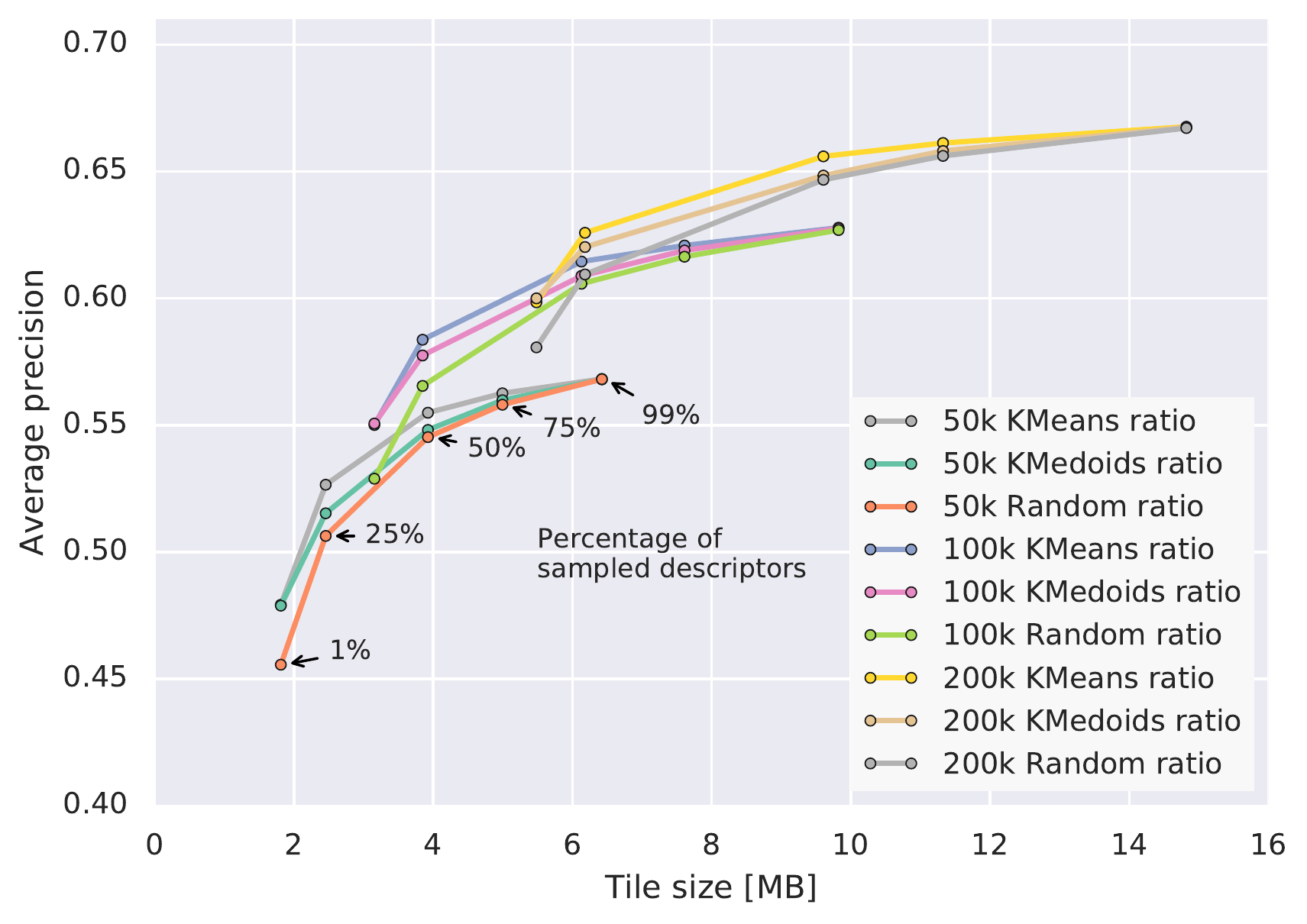}
\end{tabular}
\caption{\small Combining ILP optimization based landmark selection with appearance compression using a target fraction of the original descriptors per landmark to pick.
We compare performance on the Tokyo model using k-means or k-medoids centers with a random selection.
The minimum number of descriptors per landmark is set to 1; where the leading number denotes the different landmark budgets.}
\label{fig:chap8:ilp_ratio}
\end{figure}

\begin{figure}[]
\begin{tabular}{c}
\includegraphics[width=0.98\columnwidth]{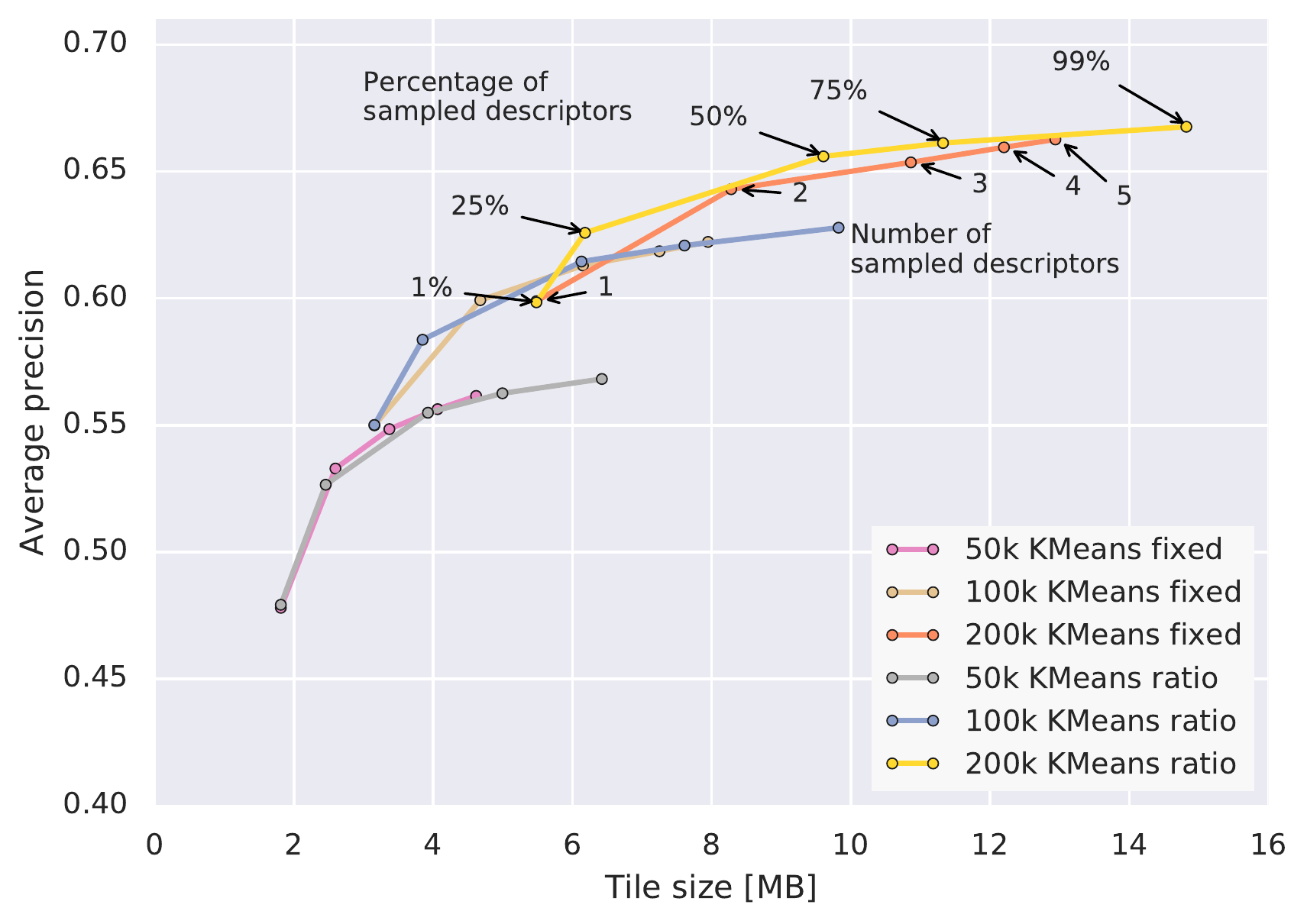}
\end{tabular}
\caption{\small Comparison of performance on the Tokyo model when picking a fixed number of descriptors per landmark vs. using a ratio of the original descriptor count as a basis.}
\label{fig:chap8:ilp_ratio_vs_fixed}
\end{figure}

Comparing a subset of variants using the `fixed count' vs. `ratio' strategies (see \reffig{fig:chap8:ilp_ratio_vs_fixed}) shows a few operating points that provide beneficial memory/performance tradeoffs: for instance 200k landmarks with a summarization to 25\% of the original descriptors.
For the majority of operating points however the difference between the two strategies is marginal.
In our system we decided to use the ILP compression picking a maximum of $200k$ landmarks per tile.
For each landmark we store $25\%$ of the original descriptors picked using kmeans.

Appearance summarization also has an important effect on the end-to-end localization performance.
Particularly in urban scenes local visual aliasing from repeated elements requires the retrieval of many neighbors during feature matching to achieve high recall on the 3D points.
Summarizing the descriptors per landmark reduces duplicate descriptors in the index and thus increases recall during matching.
%
%Previous work in the localization literature aimed at reducing the influence of such features by filtering or down-weighting them such as by \citet{camposeco2018hybrid} during compression and by \citet{sattler2016large} and \citet{sattler2017efficient} during matching.
%
%While this works for scenes which provide sufficient other information, we found performance to suffer from such weighting steps in large urban scenes where data density isn't as high as in models made from photo collections.

%------------------------------------------------------------------------------
%%%%%%%%%%%%%%%%%%%%%%%%%%%%%%%
\subsection{Appearance Compression}

\begin{figure}[]
\begin{tabular}{c}
\includegraphics[width=0.98\columnwidth]{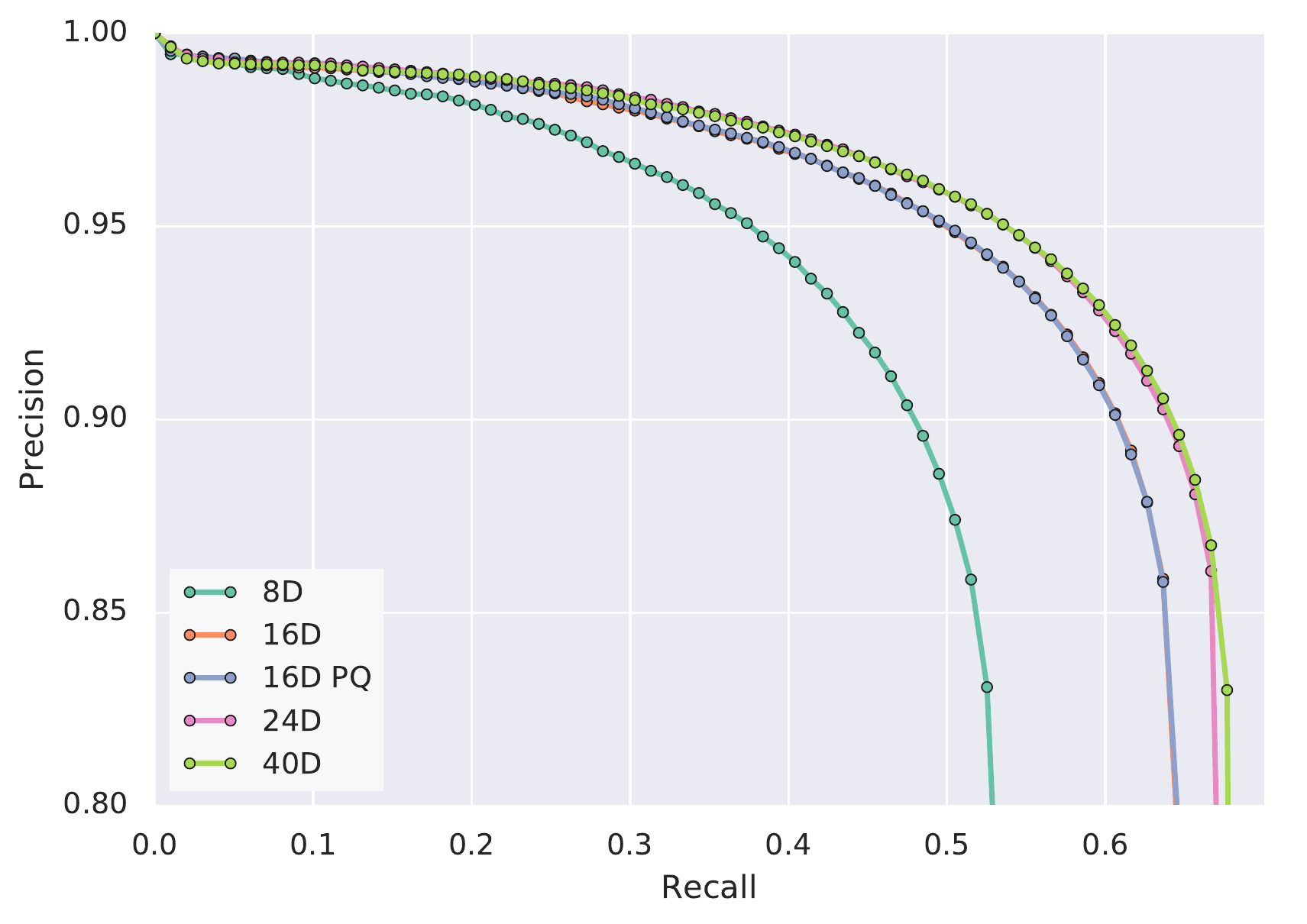}
\end{tabular}
\caption{\small Comparison of different descriptor dimensions after PCA and the impact of product quantization on average precision for the Tokyo model.}
\label{fig:chap8:dims_and_comp}
\end{figure}

After applying landmark selection and appearance summarization, we project the selected descriptors to a lower dimensional space.
The tradeoff between higher descriptor dimensions and tile size vs. average precision is shown in \reffig{fig:chap8:dims_and_comp}.
Given diminishing returns at higher dimensions we pick 16 dimensional descriptors.

As a final data reduction step we compress the descriptors using {\em Product Quantization} \citep{jegou2011PAMI}:
The 16-dimensional descriptor space is split into $M_\text{PQ}$ subspaces of equal dimensionality, \ie, each descriptor is split into $M_\text{PQ}$ parts of length $D_\text{PQ}=16 / M_\text{PQ}$.
For each subspace, a visual vocabulary with $k_\text{PQ}$ words is learned through $k$-means clustering.
These vocabularies are then used to quantize each part of a landmark descriptor individually.
A descriptor is thus represented by the indices of the closest cluster center for each of its parts.
This quantization significantly reduces storage requirements:
\Eg, when using two vocabularies ($M_\text{PQ}=2$) with $k_\text{PQ} = 256$ centers each, storing a descriptor requires only 8~bytes instead of 64~bytes.

Following \citep{jegou2011PAMI}, the squared Euclidean distance between a regular descriptor $\mathbf{d} = (\mathbf{d}_1 ~ \cdots ~ \mathbf{d}_{M_\text{PQ}})$, $\mathbf{d}_j^T \in \mathbb{R}^{D_\text{PQ}}$, and a quantized descriptor represented by a set of indices $q=(i_1, \dots, i_{M_\text{PQ}})$ is computed as
\begin{equation}
dist(d,q)^2=\sum_{j=1}^{M_\text{PQ}} \left( \mathbf{d}_j - \mathbf{c}_j(i_j) \right)^2 \enspace . \label{eq:chap8:pq_dist}
\end{equation}
Here, $\mathbf{c}_j(i_j)$ is the word corresponding to index $i_j$ in the $j^\text{th}$ vocabulary.

Decomposing the descriptor space such that each component has a similar variance reduces the quantization error of product quantization~\citep{ge2014PAMI}.
As a result, \refeq{eq:chap8:pq_dist} better approximates the true descriptor distance between the two original descriptors.
This balancing is achieved by permuting the rows of a rotation matrix that aligns the descriptor space with its principal directions as proposed by \citet{ge2014PAMI}.
Notice that this permutation does not introduce any computational overhead as the matrix is pre-computed and pre-multiplied with the projection matrix from \refsec{sec:chap8:descriptors}.

With these product quantization parameters we could not measure a performance difference between the product quantized descriptors and the 16 dimensional version (see \reffig{fig:chap8:dims_and_comp}) which suggests further potential for compression.

%------------------------------------------------------------------------------
%%%%%%%%%%%%%%%%%%%%%%%%%%%%%%%%%%%%%%%%%%%%%%%%%%%%%%%%%%%%%%%%%%%%%%%%%%%%%%%%
\section{Localization Against the Global Model}
\label{sec:chap8:model_search}
\subsection{2D-3D Descriptor Matching}
As briefly outlined in the introduction, localization is based on matching features from a query image against features stored in a pre-built model of the environment.
The first part of localization consists of establishing 2D-3D matches between the features found in the current camera and the 3D landmarks via descriptor matching \citep{li2012ECCV,li2010location,sattler2017efficient,svarm2014CVPR}.
A popular approach to accelerate the matching process is to build a kd-tree for (approximate) nearest neighbor search over the landmark descriptors \citep{li2012ECCV,svarm2014CVPR}.
While offering good search accuracy, a kd-tree is rather slow due to backtracking and irregular memory access \citep{sattler2017efficient}.
Current state-of-the-art methods for efficient large-scale localization \citep{sattler2017efficient} instead use hashing and an inverted index:
Given a fixed-size vocabulary, each feature descriptor from the current frame is assigned to its closest word.
Exhaustive search through all landmark descriptors assigned to this word then yields the nearest neighboring landmark.
This hashing approach is faster than kd-tree search and is accelerated even further through prioritization \citep{sattler2017efficient}, \ie, stopping the search once a fixed number of matches has been found.

In the following we look at different methods for accelerated nearest neighbor search besides inverted indices and also revisit the decision for the inverted multi-index we took in our previous work~\citep{Lynen:RSS2015}.

%------------------------------------------------------------------------------
%%%%%%%%%%%%%%%%%%%%%%%%%%%%%%%
\subsubsection{The inverted multi-index:}
\label{sec:chap8:model_search:imi}

Obviously, larger vocabularies are desirable as fewer descriptors will be stored for every word.
Yet, using a larger vocabulary implies higher assignment times and memory consumption.
Both of these problems can largely be circumvented by using an {\em inverted multi-index} \citep{babenko2012inverted}:
Similar to product quantization, the descriptor space is split into two parts and a visual vocabulary $\mathcal{V}_i$ containing $N_\text{imi}$ words is trained for each part.
The product of both vocabularies then defines a large vocabulary $\mathcal{V} = \mathcal{V}_1 \times \mathcal{V}_2$ containing $N_\text{imi}^2$ words.
Finding the nearest word $\omega = (\omega_1, \omega_2) \in \mathcal{V}$ for a descriptor $\mathbf{d}$ consists of finding the nearest neighboring words $\omega_1$, $\omega_2$ from the two smaller vocabularies $\mathcal{V}_1$ and $\mathcal{V}_2$, which we accelerated in our original work using a kd-tree~\citep{Lynen:RSS2015}.
Thus, using an inverted multi-index not only reduces the memory requirements but also accelerates the visual word assignments.
Each landmark descriptor is assigned to its closest word from $\mathcal{V}$.
For each feature that should be matched against the model, the $M$ nearest words from each vocabulary are found.
From the product of these two sets of words, the feature descriptor is matched against all landmark descriptors stored in the nearest words.
When using product quantization, one product quantizer is learned for each word from $\mathcal{V}_1$ and $\mathcal{V}_2$ to encode the residuals between the word and the assigned descriptors.

% \begin{figure}[t]
% \begin{tabular}{c}
% \includegraphics[width=0.98\columnwidth]{posting_list_count}
% \end{tabular}
% \caption{\small The number of filled hash buckets in the inverted multi-index for varying model sizes.}
% \label{fig:chap8:posting_list_count}
% \end{figure}

After landmark selection each sub-models of 150x150m contains around $500k$ to $1M$ descriptors.
The inverted multi-index we used previously \citep{Lynen:RSS2015}, consists of two vocabularies with 1000 words each, yielding 1 million product words.

\begin{figure}[t]
\begin{tabular}{c}
\includegraphics[width=0.98\columnwidth]{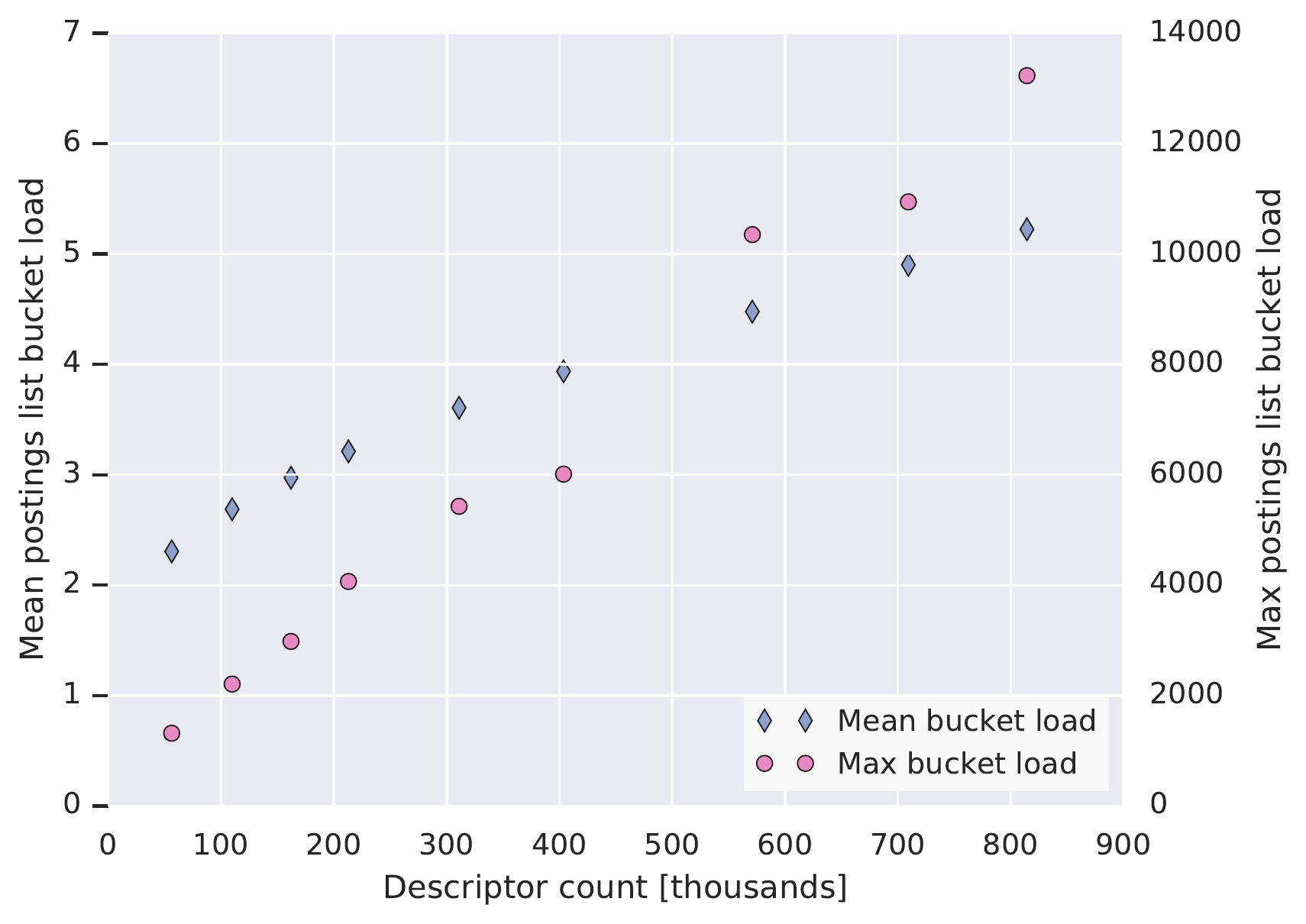}
\end{tabular}
\caption{\small The bucket load in the inverted multi-index as a function of descriptor count in tiles.
Both mean and max load increase rougly linear, but stay highly uneven over the buckets.
Naturally these lead to long tails in the processing of hash buckets (postings lists).
}
\label{fig:chap8:posting_list_bucket_load}
\end{figure}
Given that the vocabularies are trained independently, we found that only about $30\%$ of the product words in the vocabulary correspond to descriptors that are encountered in reality.
In order to obtain high recall during matching, one typically visits multiple nearest neighbors from each sub-vocabulary.
We found that between 300 and 1000 closest product words in the inverted multi-index are required for good performance.
For each word we linearly process the postings list of descriptors and compute the descriptor distance.
Because the k-means algorithm used to construct the vocabularies does not normalize the density of the descriptor space, commonly occuring descriptors are assigned to a small number of words.
We found that in our setup the postings lists of a subset of words are very long (see \reffig{fig:chap8:posting_list_bucket_load}) and cause high runtime during the query.

%------------------------------------------------------------------------------
%%%%%%%%%%%%%%%%%%%%%%%%%%%%%%%
\subsubsection{Random Grids:}

To work around the shortcomings of applying the inverted multi-index in our setting, we are interested in approaches which do not require learning a vocabulary upfront and that also handle varying density in the descriptor space better.
We found that the `Random Grids' algorithm~\citep{aiger2013random} provides these properties and is straight forward to implement and tune.
The core idea of the algorithm is to apply a random rotation and shift to the descriptor space and compute a hash-key from the transformed descriptor vector.
Descriptors from both the model and query are rotated, shifted and hashed the same way.
Whenever a query and database descriptor have the same hash-key, one computes the exact distance between the descriptor and conditionally stores the database index in the result list.
The rotation, shift and hashing functions can be picked freely (learned) such that they provide the desired runtime/quality properties.

Differences in appearance, noise and other effects mean that hash keys for corresponding descriptors can be different and the number of hash-collisions is not high enough to lead to the desired matching recall. 
Therefore the random rotations, shifts and hashing is repeated $N$ times for both database and query descriptors, which means that a database descriptor is stored in $N$ different buckets.
The different, independent rotations and shifts result in a higher number potentially colliding buckets between query and database and thus increase matching recall.

%------------------------------------------------------------------------------
%%%%%%%%%%%%%%%%%%%%%%%%%%%%%%%
\subsection{Evaluation of Matching Algorithms}

In our previous work \citep{Lynen:RSS2015}, we compared a kd-tree (used in \citep{li2012ECCV,svarm2014CVPR}), a normal inverted index (used in \citep{sattler2017efficient}), and an inverted multi-index \citep{babenko2012inverted} as techniques for establishing 2D-3D matches between query features and model points.
While the inverted multi-index provided the best runtime/quality tradeoff in our previous work, we found that the random-grids index provides even higher quality than the inverted multi-index.

For the inverted multi-index used in the following experiments, we build two vocabularies containing 1,000 words each that generate a product vocabulary with 1 million words.
At query time we use a kd-tree with approximation factor 0.1 (the error bound limiting backtracking) to identify the closest words from the product vocabulary.

The experiments focus on the tradeoff between quality and runtime by adjusting approximation parameters.
For the inverted multi-index we vary the number of closest vocabulary words visited per descriptor.
For the random-grids we vary the number of descriptors visited per grid cell.

\begin{figure}[t]
\begin{tabular}{c}
\includegraphics[width=0.98\columnwidth]{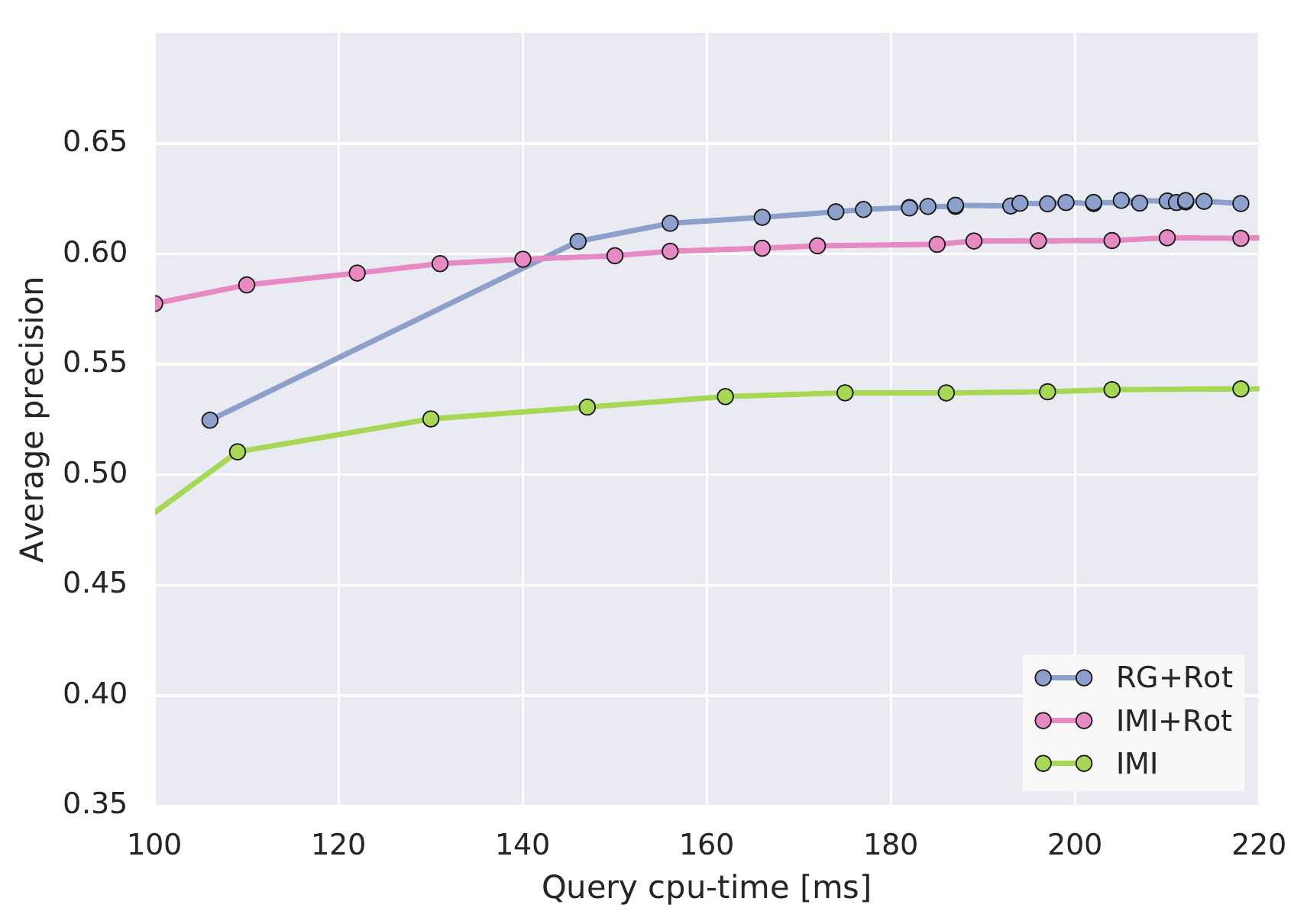}
\end{tabular}
\caption{\small Comparison of the inverted multi-index (IMI) and the random grids (RG) index using the keypoint orientation (Rot) as additional signal.
The underlying model is Tokyo compressed to 100k landmarks per tile using ILP compression.}
\label{fig:chap8:rg_vs_imi}
\end{figure}

While the random-grids approach outperform the inverted multi-index at query costs exceeding 150ms, the inverted multi-index was found to degrade more gracefully given the strategy we used for runtime restriction (see \reffig{fig:chap8:rg_vs_imi}).

Given its higher quality (at moderately increased runtime cost) we leverage the random grids algorithm with 16 dimensionsional descriptors.
We limit the runtime of random-grids by allowing a maximum of 100 descriptors per grid cell, which are picked at random.

%------------------------------------------------------------------------------
%%%%%%%%%%%%%%%%%%%%%%%%%%%%%%%
\subsection{Leveraging Priors During Matching:}
\begin{figure}[t]
\begin{tabular}{c}
\includegraphics[width=0.98\columnwidth]{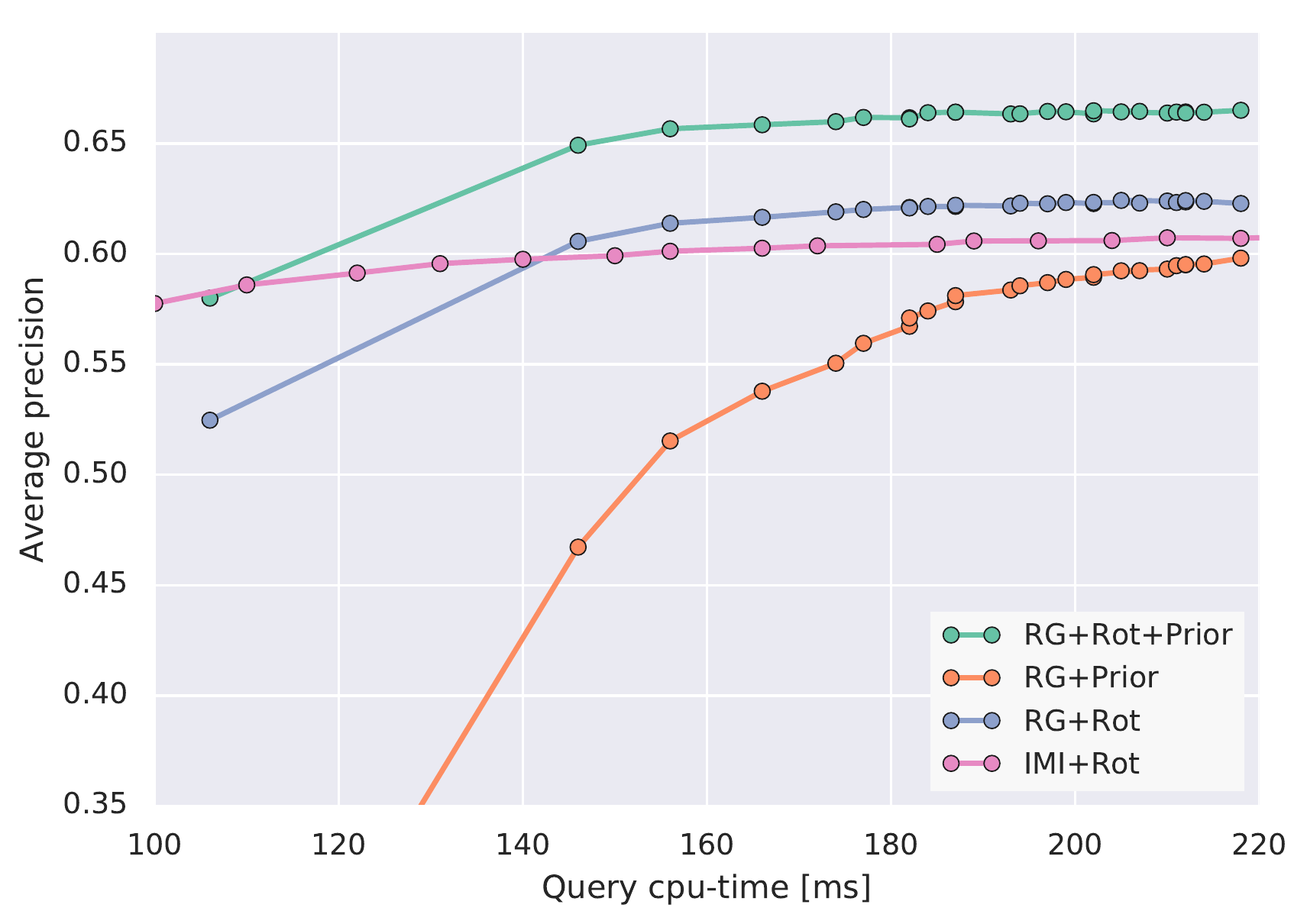}
\end{tabular}
\caption{\small Comparison of the inverted multi-index (IMI) and the random grids (RG) index using the keypoint orientation (Rot) and a location prior (Prior) as additional signals.
The underlying model is Tokyo compressed to 100k landmarks per tile using ILP compression.}
\label{fig:chap8:rg_priors}
\end{figure}

In addition to the descriptors, we store the {\em gravity rectified} keypoint orientation as proposed by \citet{kurz2011inertial}.
We extract gradient aligned CONGAS~\citep{zheng2009tour} descriptors and then change their orientation according to the gravity direction of the camera.
This method allows for faster matching (referenced as `GAFD-fast' in \citep{kurz2011inertial}) when used in the matching either as either a {\em hard} or {\em soft} filter.
Hard filtering is implemented by rejecting match candidates in the postings lists (both inverted multi-index or random-grids) if their orientation doesn't match.
Soft filtering uses the (weighted) descriptor orientation as additional descriptor dimension where it becomes part of the distance calculation used to rank neighbors.
See \reffig{fig:chap8:rg_priors} for the impact of the orientation information.

Besides using GPS information to select the models to localize against, we also leverage the information as part of the nearest neighbor search to increase matching quality.
In our use-case of outdoor robot and mobile-phone localization, priors from GPS/WiFi are typically available and accurate to between 5 and 50 meters.
Leveraging these weighted priors allows further boosting the performance for this early stage of the pipeline as shown in \reffig{fig:chap8:rg_priors}.

For keypoint orientation and GPS prior we apply weights of $0.02$ and $0.005$ respectively, found through auto-tuning using an evaluation corpus spanning multiple cities.

%------------------------------------------------------------------------------
%%%%%%%%%%%%%%%%%%%%%%%%%%%%%%%
\subsection{Range Search, kNN and Thresholding:}
\label{sec:matching_thresholds}

\begin{figure}[t]
\begin{tabular}{c}
\includegraphics[width=0.98\columnwidth]{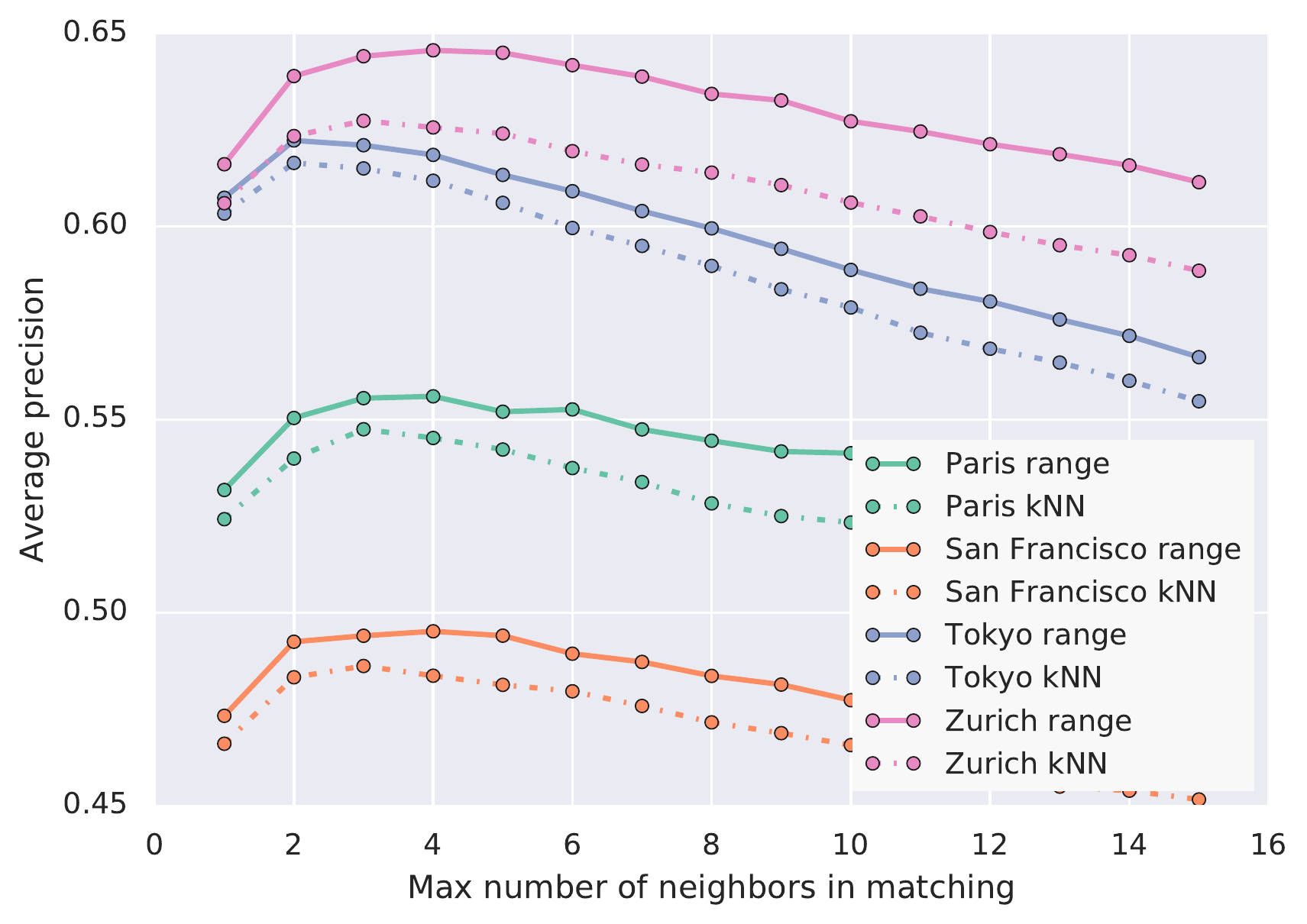}
\end{tabular}
\caption{\small Localization performance as a function of the number $k$ of retrieved database points per query keypoint
with (range) and without (kNN) thresholding on the descriptor distance.
The underlying models are compressed using the observation count compression to 500k descriptors per tile.
}
\label{fig:chap8:num_neighbors_locales}
\end{figure}

Recall of 2D-3D matches is reduced by limited descriptor distinctiveness that arises from visual aliasing and from artifacts which occur during compression.
Following previous works we thus match each keypoint from the query against $k$ neighbors from the database.
Obviously increasing $k$ leads to generally higher recall on the retrieved points but also to more outliers which negatively impact later stages.
Before pose recovery is performed, the ratio of outlier matches thus needs to be reduced to limit the runtime of RANSAC, which grows exponentially with the outlier ratio.
A set of techniques~\citep{sattler2017efficient,Zeisl2015ICCV} have been proposed to filter outlier matches.
The best choice of $k$ thus depends heavily on the choice of subsequent pipeline stages.
Having a too high outlier ratio decreases performance as shown in \reffig{fig:chap8:num_neighbors_locales} with performance peaking at $k=4$ neighbors.
As such, we use a set of outlier filters to reject wrong matches even for $k>1$.
We complement the initial filtering based on descriptor distance with more complex, geometric and co-visibility based filters described in the next section.

\begin{figure}[t]
\begin{tabular}{c}
\includegraphics[width=0.98\columnwidth]{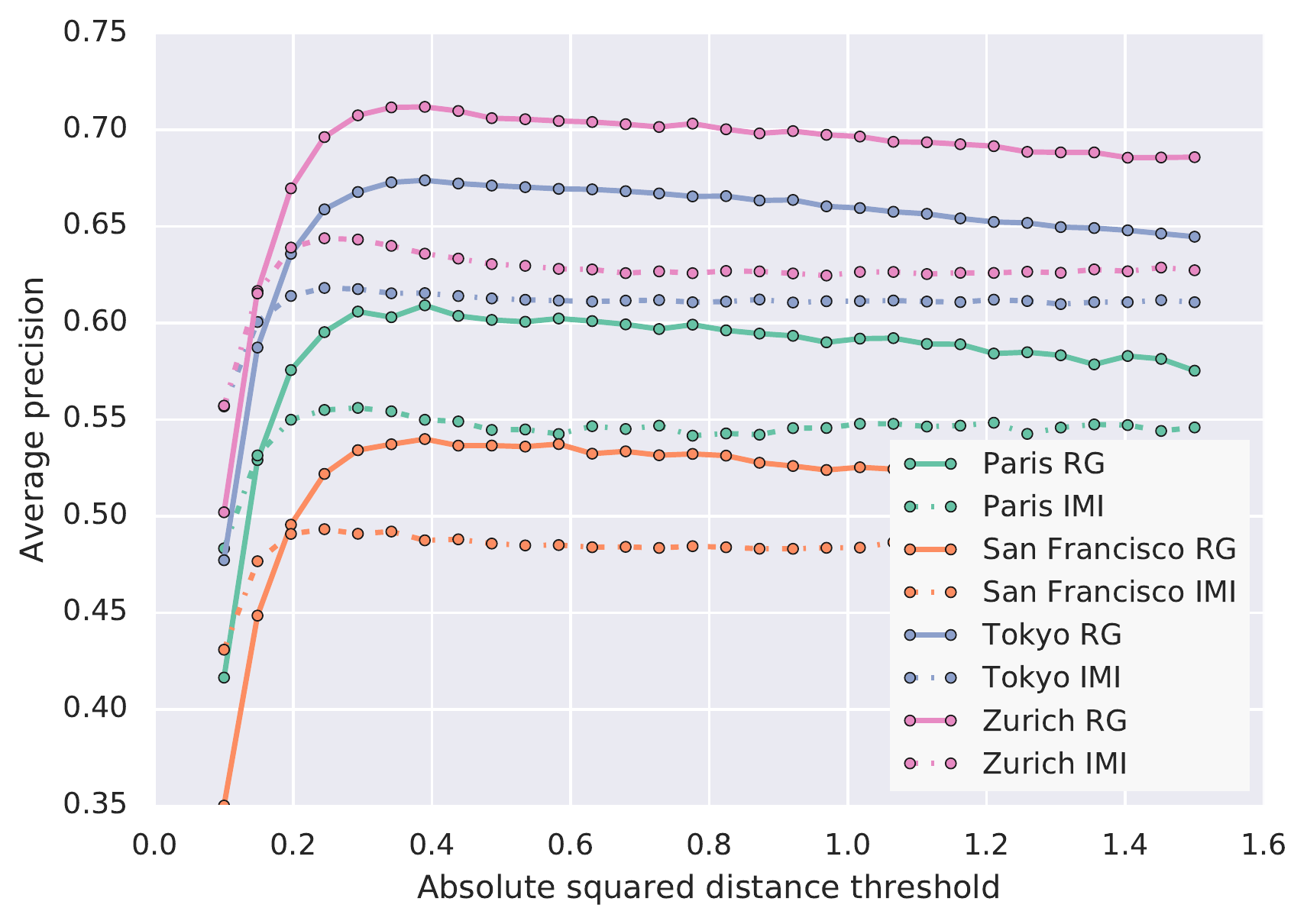}
\end{tabular}
\caption{\small The performance of Inverted Multi-Index and Random grids over a range of absolute threshold values and cities.
The optimal threshold for Random Grids is higher since it includes error thresholds for keypoint orientation and position prior.
The underlying models are compressed using the observation count compression to 500k descriptors per tile.
}
\label{fig:chap8:absolute_threshold_rg_vs_imi}
\end{figure}

We found a range search with an absolute threshold to reliably reduce outliers and improve overall performance.
The exact value of the threshold depends if the descriptor is augmented with values from the keypoint orientation and prior, but appears to be globally applicable across locales as shown in \reffig{fig:chap8:absolute_threshold_rg_vs_imi}.
While filtering on descriptor distance improves performance overall, it does not significantly change the negative impact of increasing the number $k$ of retrieved matches (see \reffig{fig:chap8:num_ransac_iterations_match_filters}).

An alternative to filtering by absolute distance is to use the {\em ratio test} proposed by \citet{Lowe:IJCV2004}.
The ratio test however turns out to be non trivial to apply when matching against large models given that it's not clear which neighbor to pick as reference:
Landmarks can have a variable number of associated descriptors and several landmarks can have near identical appearance.

\begin{figure}[t]
\begin{tabular}{c}
\includegraphics[width=0.98\columnwidth]{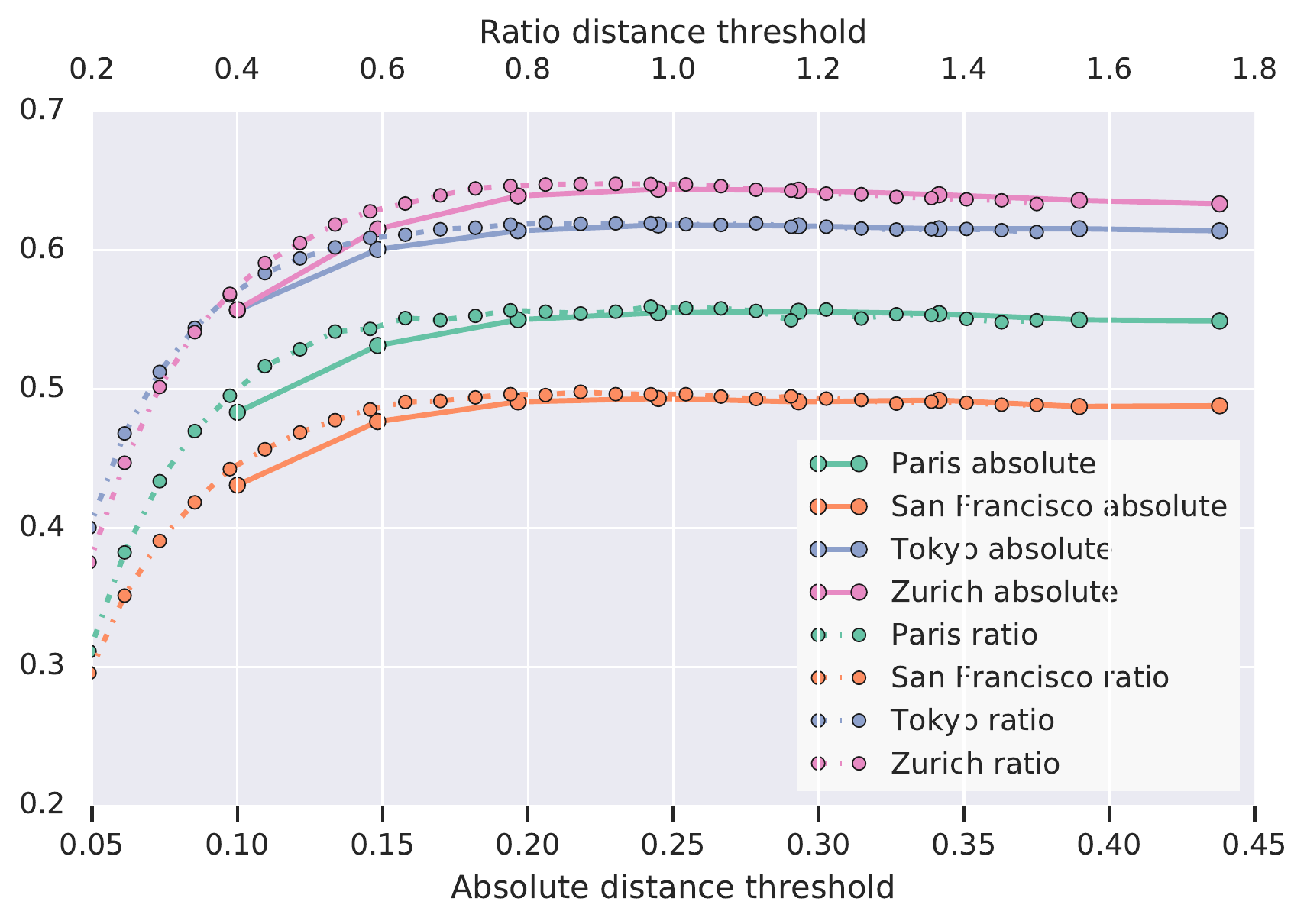}
\end{tabular}
\caption{\small The performance difference between applying an absolute threshold vs. a ratio test against close words from the vocabulary of the inverted multi-index.
The underlying models are compressed using the observation count compression to 500k descriptors per tile.
}
\label{fig:chap8:ratio_test_vs_absolute_threshold_imi}
\end{figure}

We would thus optimally obtain an independent reference (distractor) distribution to pick the descriptor distance from.
Storing this distribution and performing the search for the reference feature however incurs additional cost.
To make a fair comparison at similar runtime cost we experimented with using the vocabulary words of the inverted multi-index as reference.
The distances to the closest words provide an approximation of the local descriptor space density and thus allow deriving a distance threshold for matching that takes into account the varying density in the descriptor space.
This strategy is denoted as {\em ratio test} in \reffig{fig:chap8:ratio_test_vs_absolute_threshold_imi}.

%------------------------------------------------------------------------------
%%%%%%%%%%%%%%%%%%%%%%%%%%%%%%%
\subsection{Co-Visibility-Based and Geometric Outlier Filtering} 
\label{sec:match_filtering}
Often, fewer than $10\%$ of all features found in the current frame have a corresponding landmark \citep{sattler2017efficient} while others match to incorrect features.
Most of these wrong matches are eliminated by a threshold on the descriptor distance as discussed in the previous section.
However, some correspondences will still pass these tests, causing problems during camera pose estimation since the run-time of RANSAC increases exponentially with the outlier ratio \citep{Fischler:Bolles:ACM1981}.
An efficient and well performing technique for outlier filtering is the “pose voting” approach by \citet{Zeisl2015ICCV} which estimates for each camera pose in the map an upper bound on the inliers using geometric constraints.
The method finds the most likely camera pose by rendering surfaces that represent the geometric constraint spanned by the 2D-3D match:
Each landmark's 3D position and the angle under which is observed in the query camera span a cone in the 4D space of possible camera poses ($x$, $y$, $z$, $\kappa$).
In the original algorithm, the unknowns of the camera pose are reduced to only 2D position and the rotation ($\kappa$) around the known gravity direction.
While \citet{Zeisl2015ICCV} restricts voting to that 3D-space ($x$, $y$, $\kappa$) we vote directly in the pose 4-space showing that this results in better performance than the lower dimensional approximation.
Since the rendered surfaces lie on a two-dimensional manifold, the rendering time in 4D space is the same as for the projected 3D space used in \citet{Zeisl2015ICCV}.
After rendering the surfaces for all matches, the maximima in the voting space provide a set of likely candidate poses which are further verified as described below.

\begin{figure}[t]
\begin{tabular}{c}
\includegraphics[width=0.98\columnwidth]{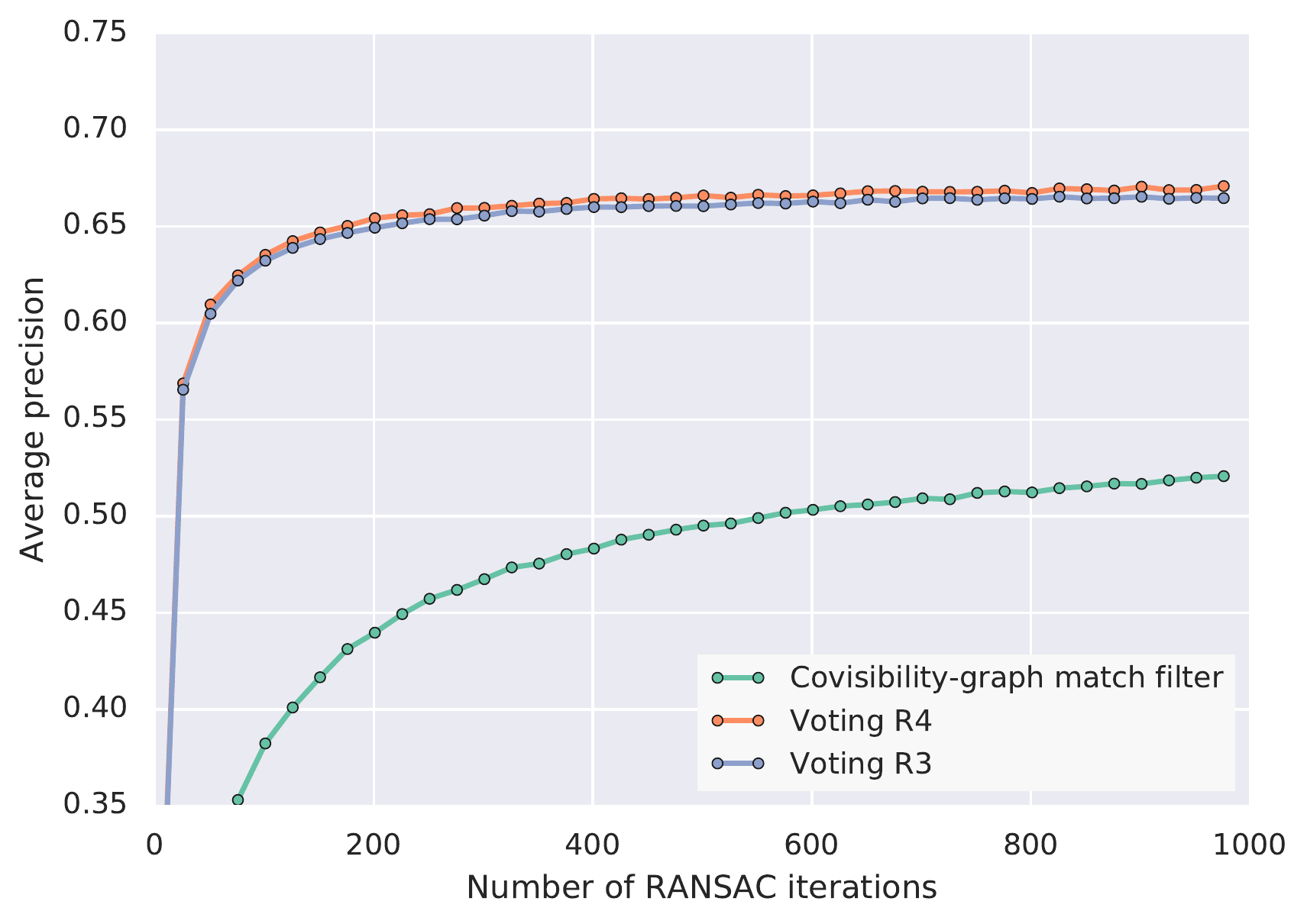}
\end{tabular}
\caption{\small Evaluation of localization performance as a function of RANSAC iterations for covisibility graph based filtering and the voting approach of \citet{Zeisl2015ICCV}.
Note the faster convergence of the voting approach at a lower count of RANSAC iterations.
The underlying models are compressed using the observation count compression to 500k descriptors per tile.
}
\label{fig:chap8:num_ransac_iterations_match_filters}
\end{figure}

This {\em covisibility filtering} approach removes matches that do not belong to the same location in the map~\citep{li2012ECCV,stumm2013probabilistic,sattler2017efficient}.
The 3D model defines a undirected, bipartite {\em visibility graph} \citep{li2010location}, where the two sets of nodes correspond to the database images and the 3D landmarks in the map.
A landmark node and a image node are connected if the landmark is visible in the corresponding database image.
The landmarks from a given set of 2D-3D matches and their corresponding database images then form a set of connected components in this visibility graph.
The covisibility filter from \citep{sattler2017efficient} simply removes all matches whose 3D point does not belong to the largest connected component.

As evaluated by \citet{lynen2017trajectory}, these algorithms however remain very sensitive to the choice of $k$ (the number of retrieved nearest neighbors per feature).
Other approaches such as the vote density based technique of \citet{lynen2017trajectory} have shown more stable performance over a wider range of values for $k$ but are computationally expensive.

\begin{figure}[t]
\begin{tabular}{c}
\includegraphics[width=0.98\columnwidth]{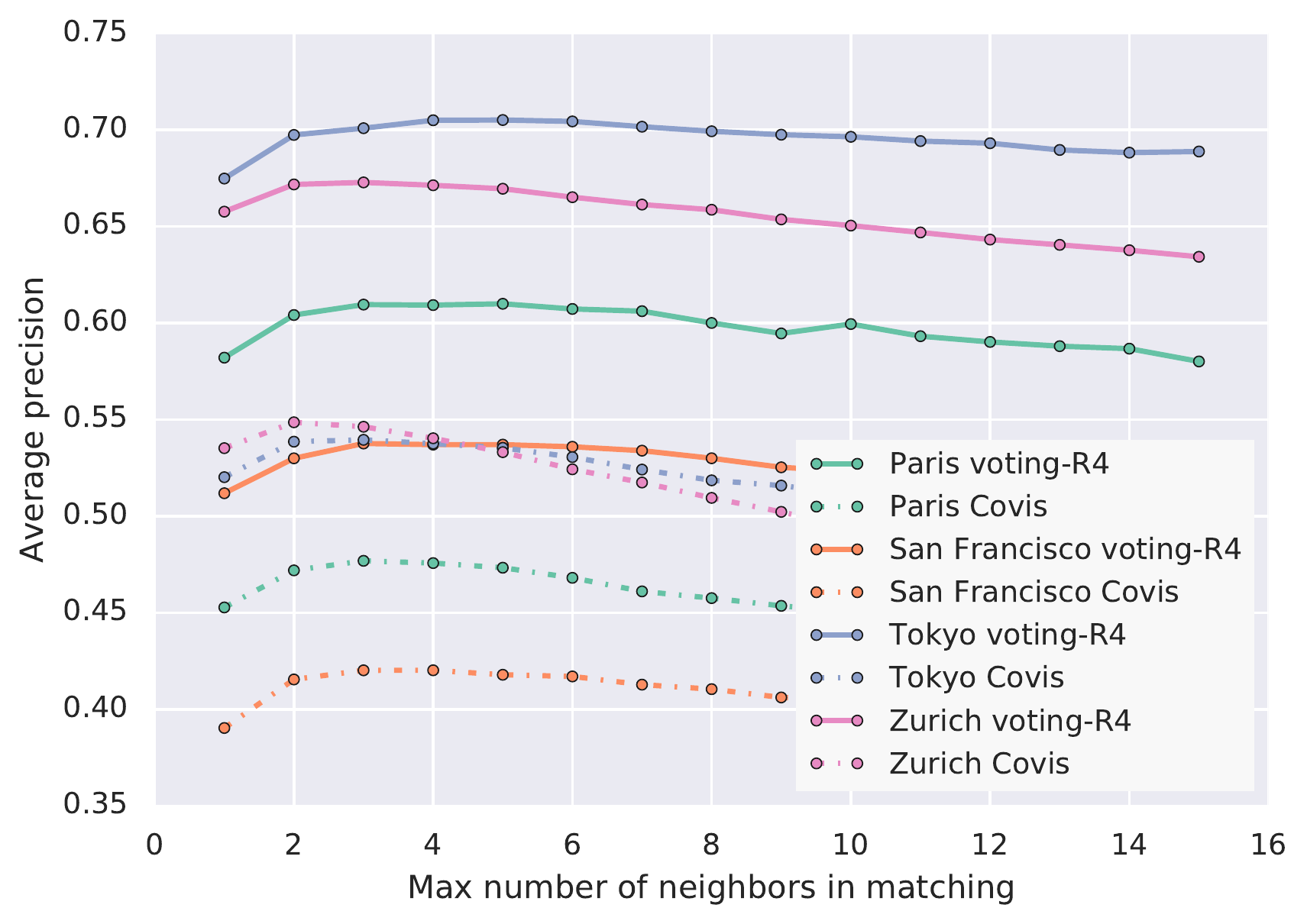}
\end{tabular}
\caption{\small Performance comparison of covisibility graph filtering and voting approaches. We found that for some sequences the approaches perform very similar (as also reported by \citet{Sattler2017CVPR}).
Over a larger number of evaluation sequences and city scale models as in our setup the sparsity varies drastically over the area and a single criterion for graph expansion is difficult to pick.
Here we found the performance of the covisibility filtering being degraded.
The underlying models are compressed using the observation count compression to 500k descriptors per tile.
}
\label{fig:chap8:voting_vs_covis_num_neighbors}
\end{figure}

We thus compare in the following voting and covisibility graph based techniques.
In \citet{Sattler2017CVPR} the covisibility graph approach outperforms the voting implementation of \citet{Zeisl2015ICCV} while for our datasets we found the opposite to be true.
It seems the optimal choice is a question of dataset and implementation (see \reffig{fig:chap8:num_ransac_iterations_match_filters} and \reffig{fig:chap8:voting_vs_covis_num_neighbors} for a comparison on city scale datasets).

%------------------------------------------------------------------------------
%%%%%%%%%%%%%%%%%%%%%%%%%%%%%%%
\subsection{Camera Pose Estimation}
\label{sec:pose_estimation}
\begin{figure}[t]
\begin{tabular}{c}
\includegraphics[width=0.98\columnwidth]{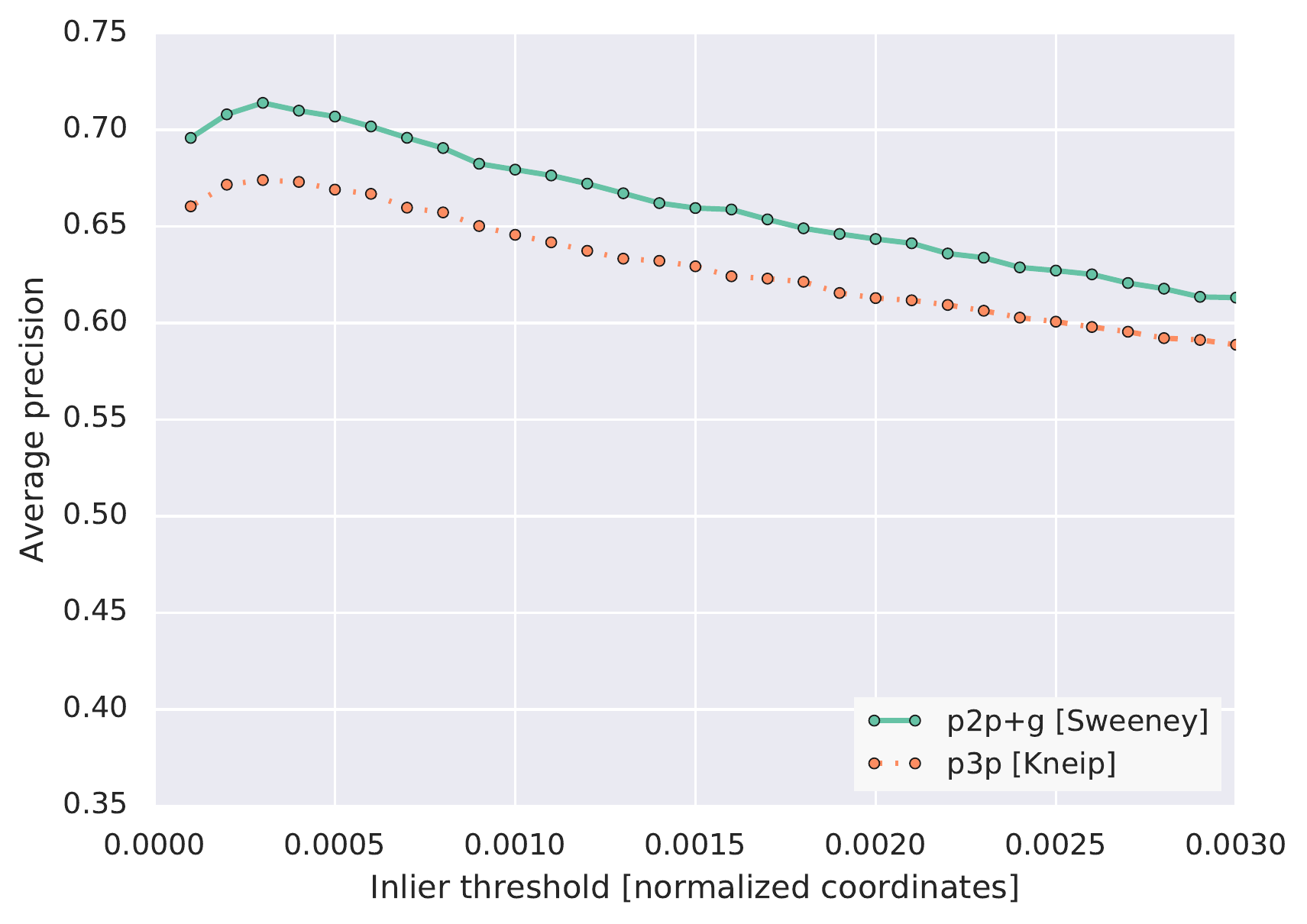}
\end{tabular}
\caption{\small The absolute pose recovery performance of p2p+g \citep{sweeney2015efficient} compared to 
P3P \citep{kneip2011novel} as a function of the inlier threshold used in RANSAC for inlier counting.
The underlying model is from Zurich and compressed using the observation count compression to 500k descriptors per tile.
}
\label{fig:chap8:pose_solver_threshold}
\end{figure}

After outlier removal we can estimate the camera pose given the matches.
Here, we follow standard techniques:
Given the knowledge of the gravity direction for both the database and the query side, we can leverage minimal solvers that exploit this additional constraint.
We found that the two-point solver proposed by \citet{sweeney2015efficient} provides superior performance to the P3P solver of \citet{kneip2011novel} as shown in \reffig{fig:chap8:pose_solver_threshold}.
To improve runtime of the pose solver we use a randomized RANSAC verification approach as proposed by \citet{matas2004randomized} and use the saved time to run additional iterations.
The final solution is refined using PnP on all inliers~\citep{hesch2011ICCV}.

%------------------------------------------------------------------------------
%%%%%%%%%%%%%%%%%%%%%%%%%%%%%%%
\subsection{Pose Candidate Refinement}
\label{sec:pose_refinement}
While increasing the number of neighbors during matching increases matching recall, the resulting increase in outliers reduces overall system performance (see \refsec{sec:matching_thresholds}).

\begin{figure}[t]
\begin{tabular}{c}
\includegraphics[width=0.98\columnwidth]{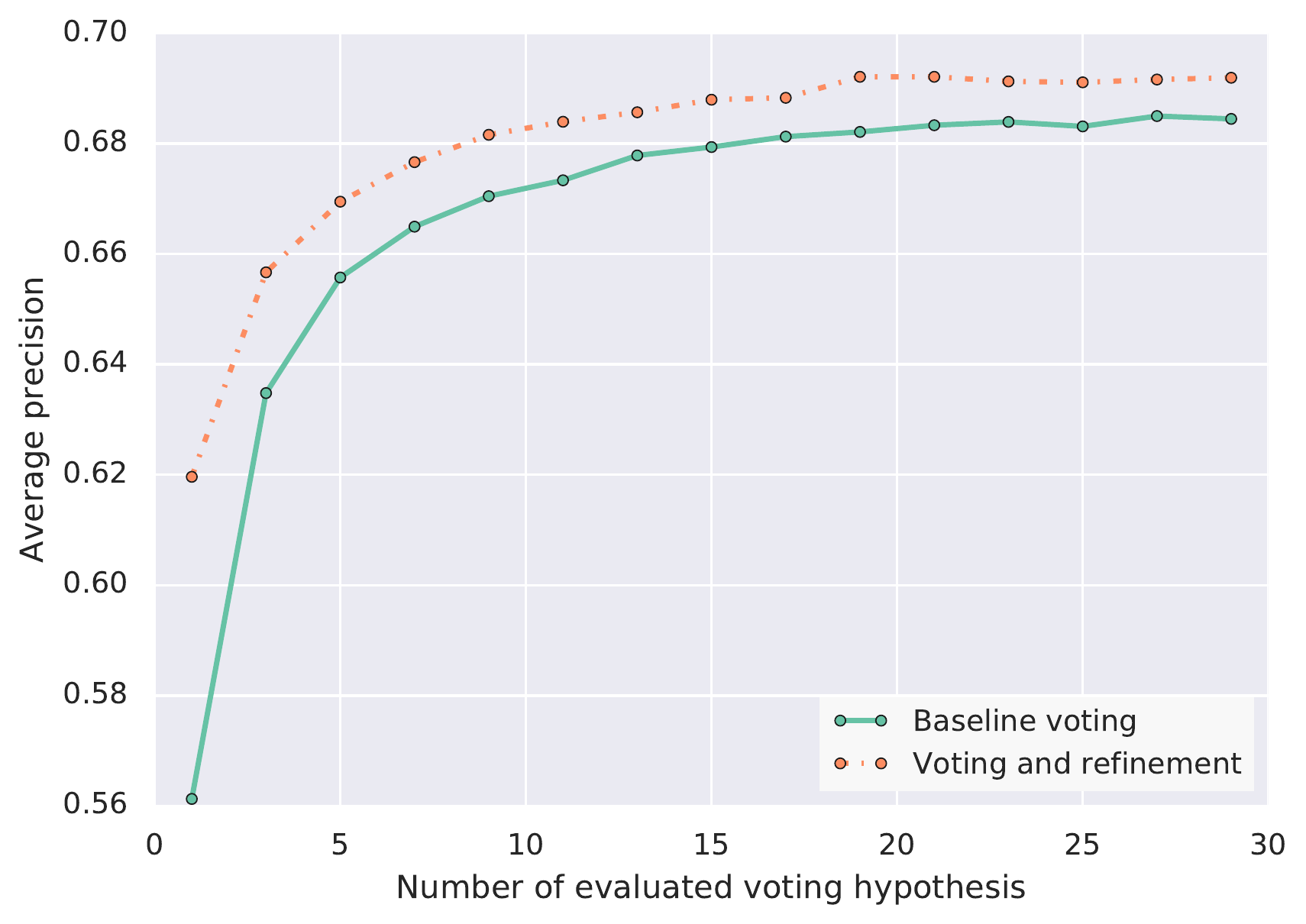}
\end{tabular}
\caption{\small Comparison of our implementation of the voting approach by \citep{Zeisl2015ICCV} with a variant using additional pose candidate refinement.
Both variants use the Random Grids matcher on descriptors augmented with keypoint orientation and position prior.
The underlying model is from Tokyo and compressed using the observation count compression to 500k descriptors per tile.
}
\label{fig:chap8:voting_neighbors_and_refinement}
\end{figure}

To recover these queries we break the localization pipeline into two hierarchical stages:
First we run the matching, filtering and pose recovery steps described in the previous sections.
We score maxima of the voting space based on the effective inlier count~\citep{irschara2009structure} and return them as pose candidates if they are deemed reliable enough.
If none of the candidates are considered good enough we run a refinement stage which breaks down the problem into smaller areas:
First we identify the relevant cameras in the model using a voting scheme similar to what was proposed by \citet{irschara2009structure,gehrig2017visual}.
We want to distinguish irrelevant regions of the map which received outlier matches (at random) from those that are relevant candidates for localization.
This differs from a covisibility filter since it includes a statistical model for the likelihood of random matches as a function of the descriptor density in the tile.
Given the assumption that irrelevant locations receive votes randomly with probability $p$, we use the random variable $X_i(q_j)$ to represent the number of matches for a keyframe $i$ given query descriptor $q_j$ which follows a binomial distribution:

\begin{equation}
  X_i(q_j) \sim \text{Bin}(n, p), n = N(q_j), p = \frac{|C_i|}{\sum_{k}{|C_k|}} \enspace ,
\end{equation}
where $x_i(q_j)$ is the number of matches for camera $i$ given query $q_j$,
$N_i := \sum_{i} x_i(q_j)$ the total number of matches for query $q_j$ over all cameras,
$|C_i|$ the number of descriptors visible from camera $i$ and
$\sum_{k}{|C_k|}$ the total number of descriptors in the model.

We expect that a location which corresponds to the true location of the query does not follow this binomial distribution.
Similar to \citet{gehrig2017visual} we formulate a Null-Hypothesis:
\textit{For an outlier pose the number of matches $x_i(q_j)$ is drawn from a binomial distribution}.
Every location in the map which has received more matches than expected given the random process
\begin{equation}
x_i(q_j) > E[X_i(q_j)] = N_i \frac{|C_i|}{\sum_{k}{|C_k|}}
\end{equation}
is thus considered a relevant location to be considered for refinement.
We weight this score with the pose prior from GPS and use it to rank candidates for refinement.
For the top $M$ locations with high scores, we match their descriptors to the descriptors from the query and run pose recovery on the resulting 2D-3D matches.

To evaluate combining pose voting with pose refinement we vary the number of candidates $P$ evaluated during the voting step and show the effect on average precision in \reffig{fig:chap8:voting_neighbors_and_refinement}.
We parallelized the pose recovery from inliers and thus limit the latency impact of increased $P$.

%------------------------------------------------------------------------------
%%%%%%%%%%%%%%%%%%%%%%%%%%%%%%%%%%%%%%%%%%%%%%%%%%%%%%%%%%%%%%%%%%%%%%%%%%%%%%%%
\section{Local Pose Tracking}

For any real-time application in robotics or augmented reality it is crucial to provide low-latency, low drift tracking of the camera pose relative to the environment.
SLAM algorithms leveraging visual and inertial data are a common choice to provide estimates relative to a {\em local} frame of reference.
By leveraging matches from the localization system one can register the local tracking against a {\em global} reference to unlock navigation and content display in the physical world.

In our system, the pose of the platform is tracked in real time using a visual-inertial sliding window estimator (implemented as an Extended Kalman Filter; EKF) with on-the-fly feature marginalization similar to the work of \citet{Mourikis:etal:TRO2009} with adaptations as proposed by \citet{hesch2014camera}.
The temporally evolving state in this estimator is given by
\begin{equation}
{\mbf x}_E = (\pose{L}{I} ~ \velocity{L}{I} ~ {\mbf b}_g ~ {\mbf b}_a ) \enspace ,
\end{equation}
where $\pose{L}{I}$ denotes the pose of the platform as the coordinate transformation\footnote{Internally represented as a vector for the translation and for the rotation, a unit-quaternion in JPL notation \citep{trawny2005indirect}.} of the IMU frame of reference \wrt the local SLAM frame of reference.
The translational velocity estimate of the IMU frame \wrt the local SLAM frame is denoted as $\velocity{L}{I}$.
${\mbf b}_g$, ${\mbf b}_a \in \mathbb{R}^3$ denote the estimate of the time varying gyroscope and accelerometer bias, modeled as random walk processes driven by the zero-mean, white, Gaussian noise vectors ${\mbf n}_{bg}$ and ${\mbf n}_{ba}$.

Besides the evolving state ${\mbf x}_E$, the full estimated state $\hat {\mbf x}_k$ at time $k$ also includes the position and orientation of $N$ cameras which form the sliding window of poses \citep{Mourikis:etal:TRO2009}:
\begin{equation}
\hat {\mbf x}_k = (\hat{\mbf x}_E ~ \pose{L}{C_1} ~ \cdots ~ \pose{L}{C_N} ) \enspace .
\end{equation}
Here, $\pose{L}{C_i}$, $i = 1 \ldots N$, denote the estimates of the pose of the $i$th camera.

Using measurements from the IMU increases robustness and accuracy while providing a metric pose estimate.
At the same time, the proper marginalization of past measurements \citep{dong2011motion,sibley2010sliding,nerurkarc:etal:RSS2013,leutenegger2014keyframe} is key to obtain a smooth pose estimate.
This is particularly relevant when including measurements to the global model which are often not in perfect agreement with the locally observed structure, \eg, due to moving objects during model creation or drift in the local pose estimates.

%------------------------------------------------------------------------------
%%%%%%%%%%%%%%%%%%%%%%%%%%%%%%%
\subsection{Global Updates to the Local State Estimation}

In order to boot-strap the localization system, descriptors from the cameras of the local SLAM system are matched against the model as described in \refsec{sec:chap8:model_search}.
Once an estimate of the pose $\pose{G}{C}$ of the camera \wrt the global model is available, the frame of reference of the local SLAM system is aligned with the global map using the relative transformation from global model to local SLAM frame of reference $\pose{G}{L}$:
\begin{eqnarray}
\pose{G}{L} & = & \pose{G}{C} \oplus \pose{C}{I} \oplus \pose{I}{L} \\
\pose{I}{L} & = & \pose{L}{I}^{-1}.
\end{eqnarray}
Here, $\otimes$ denotes the transformation composition operator.
The computed transformation $\pose{G}{L}$ is subsequently integrated into the state of the SLAM system and continuously refined using 2D-3D matches from the localization system.
Once the alignment of local SLAM and global map is established, we use the SLAM system's state estimate to filter subsequent measurements using a Mahalanobis distance check.
We found this to further improve the performance of the system even though most outliers are already filtered by the RANSAC step in the pose estimation during localization.

In the algorithm proposed in this paper, every associated 2D-3D match provides a measurement of the form
\begin{equation}
{\mbs z}_i^{(j)}=\frac{1}{^{C_i}z_j} \begin{bmatrix} ^{C_i}x_j \\ ^{C_i}y_j \end{bmatrix} + {\mbs n}_i^{(j)} \enspace ,
\end{equation}
where $[^{C_i}x_j~^{C_i}y_j~^{C_i}z_j]^T =~^{C_i}{\mbs p}_j$ denotes the position of the $j$th 3D point expressed in the frame of reference of camera $i$.
To obtain the residual for updating the EKF, we express the expected measurement $\hat{\mbs z}_i^{(j)}$ as a function $h$ of the state estimate $\hat {\mbf x}_k$ and the position of the landmark $^G\mbs{p}_\ell$ expressed in the global frame of reference:
\begin{equation}
{\mbs r}_i^{(j)} = {\mbs z}_i^{(j)} - \hat{\mbs z}_i^{(j)} = {\mbs z}_i^{(j)} - h(^G{\mbs p}_\ell, \pose{G}{C_i}) \enspace.
\end{equation}
By linearizing this expression around the state estimate we obtain (see \citep{Mourikis:etal:TRO2009}, Eq. (29) for details):
\begin{equation}\label{eq:chap8:residual}
{\mbs r}_i^{(j)} \simeq {\mbs H}_{GL_i}^{(j)}({\mbs x}_i - \hat{\mbs x}_i) + {\mbs n}^{(j)} \enspace .
\end{equation}
Here, ${\mbs H}_{GL_i}^{(j)}$ denotes the global landmark measurement Jacobian with non-zero blocks for the pose of camera $i$.

When querying the map, it is not unusual to retrieve hundreds of matches from the image to the global map.
To reduce the computational complexity of updating the estimator, all the residuals and Jacobians ${\mbs r} = {\mbs H}({\mbs x} - \hat {\mbs x}) + {\mbs n}$ are stacked to apply measurement compression \citep{Mourikis:etal:TRO2009}.
More specifically, we apply a QR-decomposition to ${\mbs H}$:
\begin{equation}
{\mbs H} = \begin{bmatrix}{\mbs V}_1 ~ {\mbs V}_2\end{bmatrix}\begin{bmatrix}{\mbs T}_H \\ {\mbs 0}\end{bmatrix} \enspace ,
\end{equation}
where ${\mbs T}_H$ is an upper triangular matrix and ${\mbs V}_1$, ${\mbs V}_2$ are unitary matrices with columns forming the basis for the range and the nullspace of ${\mbs H}$, respectively.
This operation projects the residual on the basis vectors of the range of ${\mbs H}$, which means that ${\mbs V}_1$ extracts all the information contained in the measurements.
The residual from \refeq{eq:chap8:residual} can then be rewritten as
\begin{equation} \label{eq:chap8:meas_compression}
\begin{bmatrix}{\mbs V}_1^T {\mbs r} \\ {\mbs V}_2^T {\mbs r}\end{bmatrix} = \begin{bmatrix}{\mbs T}_H \\ {\mbs 0}\end{bmatrix}({\mbs x} - \hat {\mbs x}) + \begin{bmatrix}{\mbs V}_1^T {\mbs n} \\ {\mbs V}_2^T {\mbs n}\end{bmatrix} \enspace .
\end{equation}
After discarding ${\mbs V}_2^T {\mbs r}$ in \refeq{eq:chap8:meas_compression} since it only contains noise, we obtain the compressed residual formulation
\begin{equation}
{\mbs r}_n = {\mbs V}_1^T {\mbs r} = {\mbs T}_H ({\mbs x} - \hat {\mbs x}) + {\mbs n}_n \text{~with~} {\mbs n}_n = {\mbs V}_1^T {\mbs n} \enspace.
\end{equation}
Using Givens rotations, the residual ${\mbs r}_n$ and the upper triangular matrix ${\mbs T}_H$ for $L$ matches can be computed efficiently in $\mathcal{O}((6N)^2L)$.

The MSCKF algorithm \citep{Mourikis:etal:TRO2009} processes a feature track as soon as the track is broken or the entire window of cameras is spanned by the track.
In our local SLAM system, these completed tracks are used to triangulate landmarks and the resulting 3D points are then used to update the estimator.
When using the same visual features for frame-to-frame tracking and global localization, we can use a match between the 3D model and a feature from a single frame to identify the corresponding feature track.
This information can now be used to form a constraint to the 3D model that involves all cameras that are spanned by the corresponding feature track.
We found that forming a constraint which involves all key-frames which are part of the track gives a lower tracking error than performing single camera updates.
To avoid double counting information, care must be taken that feature measurements are used to either formulate a constraint in the local SLAM {\em or} to the global 3D map, but not both.
Given the information content of the global map, measurements to the global 3D map are preferred; where additional constraints on the updates can ensure estimator consistency~\citep{hesch2012consistency}.

In order to keep memory requirements bounded, we do not store the covariance matrix for the landmarks in the map, but instead inflate the measurement noise covariance during the update assuming the landmark position errors are uncorrelated between observations.
While being approximate and thus not the most accurate, the strategy allows us to keep the storage cost for a map limited.
A better filtering formulation named `Cholesky-Schmidt-Kalman filter' was recently proposed by \citet{dutoit2017consistent}, which allows for consistent updates and thus showed a reduction in pose error by more than $25\%$.
The storage cost for this approach is only linear in the number of map landmarks, but increases the computational cost for the update by about 20x.
In this paper, we thus continue to assume the map to be perfect and inflate the measurement noise of the update accordingly.

%------------------------------------------------------------------------------
%%%%%%%%%%%%%%%%%%%%%%%%%%%%%%%
\subsection{Evaluation of the Pose Fusion}
\label{sec:chap8:experiments:pose_tracking}

Next, we evaluate the pose tracking quality achieved with our state estimator that directly includes global 2D-3D matches as EKF updates.
We are interested in the position and orientation accuracy of our system, as well as the smoothness of the computed trajectories, and the time required for the state updates.
We compare our system to the algorithm proposed by \citet{middelberg2014scalable} which performs local SLAM using a sliding window BA which optimizes only local parameters and keeps the global model fixed.

Both \citet{middelberg2014scalable} and \citet{ventura2014global} first compute an initial alignment using either the camera poses returned by the server or the global landmark positions.
For all following frames that are sent to the server, they then optimize the alignment by including the global 2D-3D matches into the bundle adjustment of the local map.
To limit the computational complexity, they perform a windowed version of SLAM based on a limited number of cameras.
Cameras which are furthest away from the current pose \citep{middelberg2014scalable} or oldest \citep{ventura2014global} are discarded together with their constraints to the local and global model.
Discarding (instead of properly marginalizing) measurements however has been shown to lead to suboptimal estimation performance \citep{dong2011motion,sibley2010sliding,leutenegger2014keyframe}.
The removal of constraints from the optimization also leads to discontinuities in the resulting pose estimate as the minimum of the cost function changes.
To improve the performance of our reference implementation and allow a fair comparison we included IMU constraints in the bundle adjustment and use a non-linear solver that exploits the structure of the problem.
\reftab{tab:EKF_vs_BA_timing} compares our method against the approach of \citep{middelberg2014scalable} in which both estimators are fed with the exact same data and the same constraints to the map.
For every camera frame, we evaluate the error between the estimated pose and the ground truth as the Euclidean distance between the positions and the disparity angle in orientation.
As evident from the table, our EKF-based approach achieves a positional accuracy similar to the best-performing bundle adjustment approach while offering a more accurate estimate of the camera orientation.
At the same time, our approach is more than one order of magnitude faster than the most efficient bundle adjustment based variant.
While BA is typically run in a separate thread to avoid blocking pose tracking \citep{middelberg2014scalable}, the proposed EKF-based estimator is efficient enough to be directly run on each frame.

\begin{table}[]
\caption{Comparing the proposed EKF-based estimator update with sliding window bundle adjustment (BA): We report the mean time {\em t-up} required to update the estimator based on the global 2D-3D matches and the mean position $||\bar{p}_{err}||$ and orientation error $||\bar{\theta}_{err}||$ of each method (incl. std-dev.). For sliding window bundle adjustment, we experiment with using different numbers of cameras in the window (first number) and different numbers of bundle adjustment iterations (second number).}
\label{tab:EKF_vs_BA_timing}
\centering
\begin{tabular}{| l | c | c | c | c | r  |}
    \hline
 & t-up [ms] &  $||\bar{p}_{err}||$ [m] &  $||\bar{\theta}_{err}||$ [deg]\\ \hline
EKF & $2.9\pm1.5$ & $0.17\pm0.12$ & $0.32\pm0.16$\\ \hline
BA-10-10 & $163.0\pm43.0$ & $0.13\pm0.15$ & $0.41\pm0.17$  \\ \hline
BA-10-5 & $138.8\pm36.5$ & $0.12\pm0.14$ & $0.58\pm0.15$ \\ \hline
BA-5-10 & $100.3\pm31.9$ & $0.11\pm0.13$ & $0.53\pm0.15$  \\ \hline
BA-5-5 & $77.6\pm31.6$ & $0.12\pm0.14$ & $0.57\pm0.10$ \\ \hline
BA-5-2 & $38.6\pm14.4$ & $0.14\pm0.16$ & $0.54\pm0.16$\\
\hline
  \end{tabular}
\end{table}

\begin{figure}[]
\begin{tabular}{c}
\includegraphics[width=0.98\columnwidth]{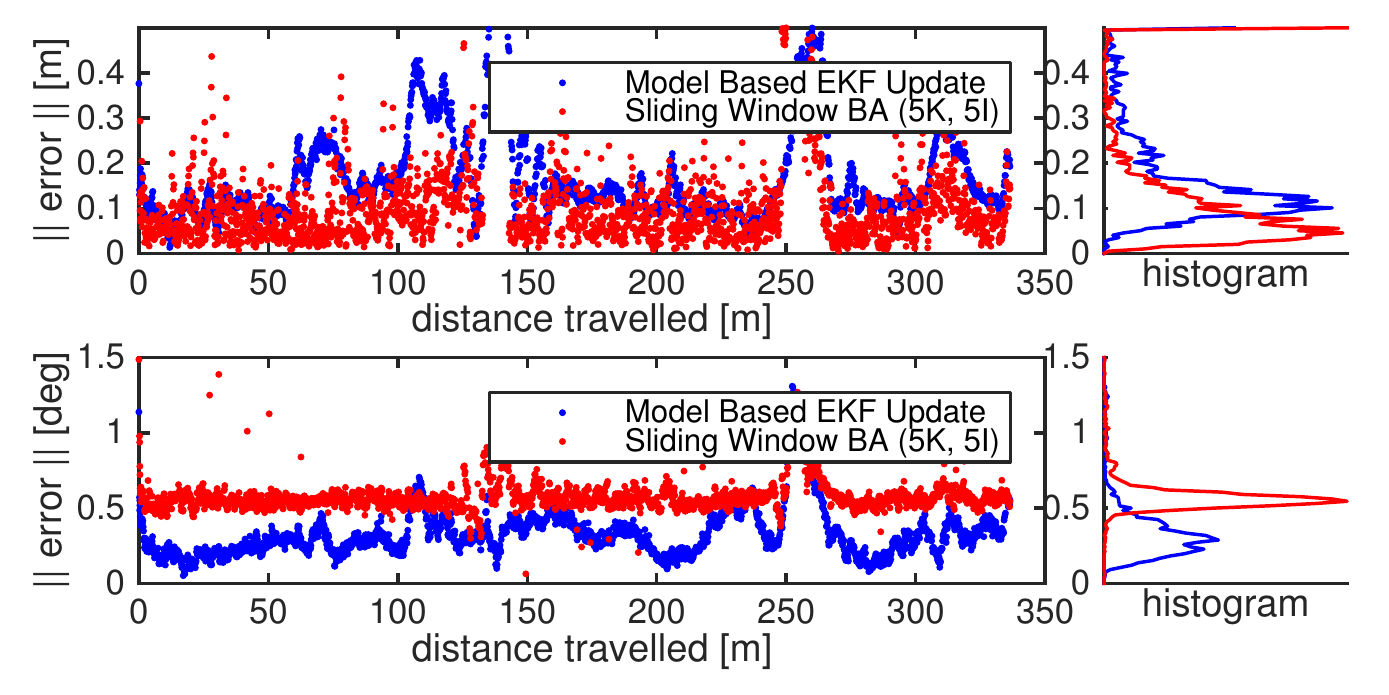} \\
\includegraphics[width=0.98\columnwidth]{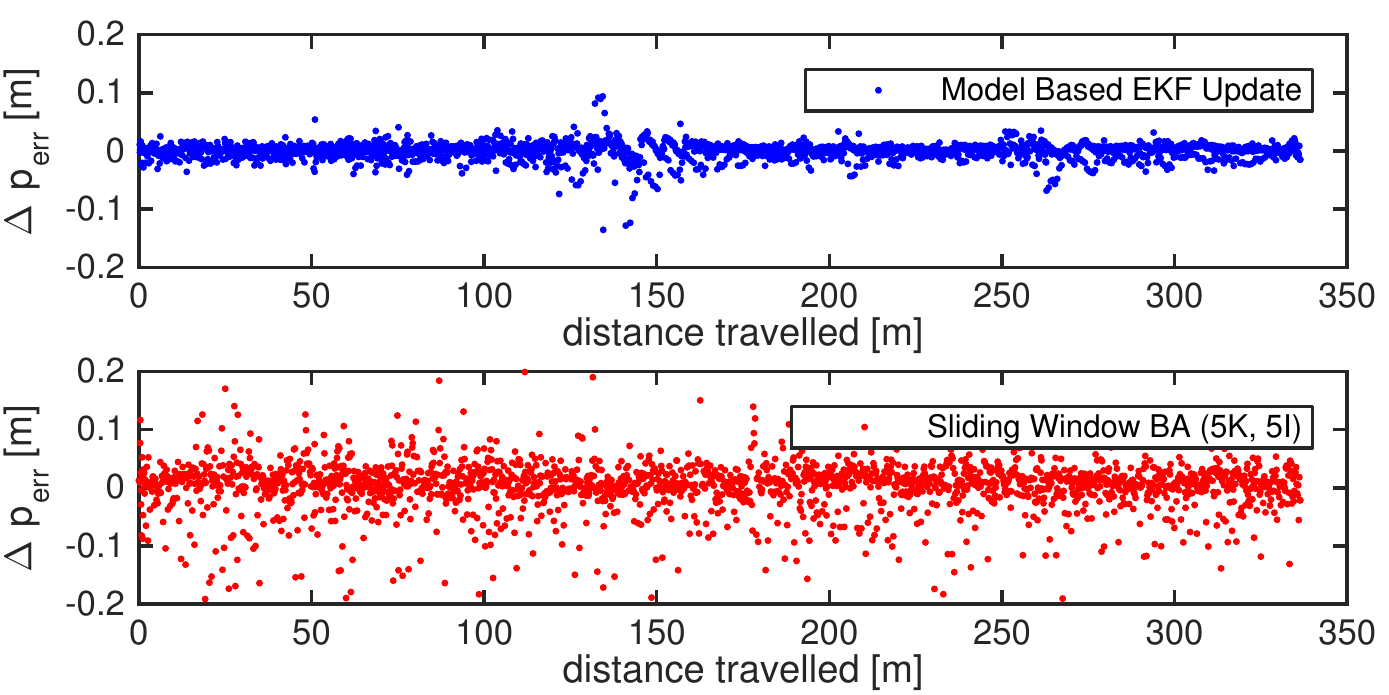}
\end{tabular}
\caption{\small Compared to a sliding window bundle adjustment approach similar to \citep{middelberg2014scalable}, the proposed direct inclusion of the global 2D-3D matches into EKF gives significantly smoother trajectories as evident from the difference in the pose error between subsequent frames.}
\label{fig:pose_tracking}
\end{figure}
 
\reffig{fig:pose_tracking} shows that even though both approaches offer a similar mean pose accuracy, our estimator achieves a much better temporal consistency.
We capture this measure by comparing the change in the error between ground truth and estimate for position and orientation.
Avoiding discontinuities in pose tracking is a vital property for obstacle avoidance and robot control, where large pose jitter can cause problems when determining a path that prevents a collision.
Both the runtime and accuracy metrics underline the superiority of a pose fusion approach that properly marginalizes the global constraints.

%------------------------------------------------------------------------------
%%%%%%%%%%%%%%%%%%%%%%%%%%%%%%%%%%%%%%%%%%%%%%%%%%%%%%%%%%%%%%%%%%%%%%%%%%%%%%%
\section{Conclusion} 
In this paper, we have presented a system for visual localization at scale.
We provide details about map compression and localization algorithms that we found essential for scalable server-side deployment.

We leverage map compression techniques from the Computer Vision community and extend them with descriptor summarization to gain another 2x-4x reduction in memory footprint.
With our detailed analysis of localization performance as function of map-compression algorithms we aim to provide guidance for parameter choices to the community.
We also show the impact of various parameters in the localization stack including novel insights into outlier filtering strategies that allow queries with latencies in the 200ms range.
Through the addition of an improved descriptor matching algorithm, position prior and a candidate refinement step we found the system to outperform other approaches when working with large scale models.
We evaluate a proof-of-concept implementation of the system across four cities from three continents by querying 2.5 million images collected with smart-phone cameras.
The scale of models and number of queries surpasses previous evaluations by several orders of magnitude and hopefully demonstrates the wide applicability of the parameter choices.

%------------------------------------------------------------------------------
%%%%%%%%%%%%%%%%%%%%%%%%%%%%%%%%%%%%%%%%%%%%%%%%%%%%%%%%%%%%%%%%%%%%%%%%%%%%%%%%
%\begin{acks}
%The research leading to these results has received funding from Google.
%\end{acks}

\bibliographystyle{plainnat}
\bibliography{robotvision}

\end{document}